\documentclass[preprint,authoryear,3p]{elsarticle}

\input{epsf}

\usepackage{bold-extra}
\usepackage[english]{babel}
\usepackage[latin1]{inputenc}
\usepackage{natbib}
\usepackage{amssymb}
\usepackage{amsmath}  

\usepackage{xcolor}    
\usepackage{moreverb} 
\usepackage{rotating}
\usepackage{tabularx}
\usepackage{stmaryrd}

\usepackage{pdfpages}
\usepackage{morefloats}
\usepackage{graphics}
\usepackage{bbm}
\usepackage{epsfig}
\usepackage{mathrsfs}
\usepackage{multirow}
\usepackage{amsthm}
\usepackage{graphicx} 

\usepackage{geometry}
\usepackage{microtype}

\newtheorem{df}{Definition}

\newproof{pf}{Proof}

\newcommand{\abs}[1]{\mid #1 \mid}

\newcommand{\accuracy}[2]{\mbox{\em acc}^#1_#2}
\newcommand{\heads}[1]{\mbox{\em heads}_#1}
\newcommand{\pheads}[2]{\mbox{\em heads-pred}^#1_#2}

\newcommand{\nemesid}{{\sc n}\textnormal{e}{\sc m}\textnormal{e}{\sc s}\textnormal{i}{\sc d}}

\journal{Neural Networks}

\begin{document}

\begin{frontmatter}

\title{Early Stopping by Correlating Online Indicators in Neural Networks}  
  
\author[UVigo]{Manuel Vilares Ferro\corref{cor}}
\cortext[cor]{Corresponding author: tel. +34 988 387280, fax +34 988 387001.}
\ead{vilares@uvigo.es}
\author[UVigo]{Yerai Doval Mosquera}
\ead{yerai.doval@uvigo.es}
\author[UVigo]{Francisco J. Ribadas Pena}
\ead{ribadas@uvigo.es}
\author[UVigo]{V\'ictor M. Darriba Bilbao}
\ead{darriba@uvigo.es}

\address[UVigo]{Department of Computer Science, University of Vigo \\ Campus As Lagoas s/n, 32004 -- Ourense, Spain}

\begin{abstract}
  In order to minimize the generalization error in neural networks, a
  novel technique to identify overfitting phenomena when training the
  learner is formally introduced. This enables support of a reliable
  and trustworthy early stopping condition, thus improving the
  predictive power of that type of modeling. Our proposal exploits the
  correlation over time in a collection of online indicators, namely
  characteristic functions for indicating if a set of hypotheses are
  met, associated with a range of independent stopping conditions
  built from a canary judgement to evaluate the presence of
  overfitting. That way, we provide a formal basis for decision making
  in terms of interrupting the learning process.

  As opposed to previous approaches focused on a single criterion, we
  take advantage of subsidiarities between independent assessments,
  thus seeking both a wider operating range and greater diagnostic
  reliability. With a view to illustrating the effectiveness of the
  halting condition described, we choose to work in the sphere of
  natural language processing, an operational \textit{continuum}
  increasingly based on machine learning. As a case study, we focus on
  parser generation, one of the most demanding and complex tasks in
  the domain. The selection of cross-validation as a canary function
  enables an actual comparison with the most representative early
  stopping conditions based on overfitting identification, pointing to
  a promising start toward an optimal bias and variance control.
\end{abstract}


\end{frontmatter}

\section{Introduction}
\label{section-introduction}

A \textit{neural network} ({\sc
  nn})~\citep{Hinton:1989:CLP:74807.74812,Nilsson:1990:MFL:87765,Rumelhart:1986:LIR:104279.104293}
is a \textit{machine learning} ({\sc ml}) system that mimics the
behavior of the human brain~\citep{McCulloch:1988:LCI:65669.104377}
for capturing underlying relationships in a set of data from a
collection of training examples. Referred to as
\textit{generalization}, this ability to correctly apply the knowledge
gained to new situations is a common reference of how accurately a
{\sc ml} algorithm is capable of interacting in practice, and is usually
measured by the \textit{generalization error}. Looking for a suitable
representation and better use of data, a {\sc nn} can integrate
multiple layers to progressively extract higher-level features from
the raw input. We then talk about \textit{deep learning} ({\sc dl})
architectures~\citep{Bengio:2013:RLR:2498740.2498889,Bengio-etal-2015,Schmidhuber:2015:DLN:2947706.2947734},
an evolution from the shallow one with profound implications for the
expressiveness of the models generated~\citep{Bianchini-etal-2014},
but that, in any case, requires generalization errors to be minimized if
we want to exploit their potential to the full. At this point, a key
factor to consider is the avoidance of \textit{overfitting} phenomena.
Also known as \textit{overlearning} or \textit{overtraining}, this
concept refers to the production of an analysis by a learner that
corresponds too closely to the training set. When that happens, the
resulting model may fail to reliably predict future observations.


Overfitting arises in models with low \textit{bias} and high
\textit{variance}, i.e., when the learner focuses on the detail and/or
noise in the training data, often as a result of its structural
complexity. That way, in generating large parameter spaces of possibly
billions of trainable items~\citep{Goodfellow:2016}, {\sc
  nn}{\footnotesize s} are prone to overfitting
behavior~\citep{Geman:1992:NNB:148061.148062}. Since neither bias nor
variance are monotonic even in simple
cases~\citep{Baldi:1991:TEG:1350931.1350941}, it is not always
possible to establish when an increase in generalization error signals
true overfitting, which explains both the need to simplify the
hypotheses for the generated models and to have robust diagnostic
strategies. Following~\cite{Prechelt97automaticearly}, we then talk
about \textit{regularization} in the first case and \textit{early
  stopping} in the second one.

There are basically two ways of applying regularization: reducing the
number of dimensions of the parameter space --usually the connection
weights in the network-- or reducing the effective size of each
dimension. In the latter case, \textit{weight decay}~\citep{krogh:92}
has become a reference method, while in the first one a number of
different approaches have been proposed, including \textit{greedy
  constructive learning}~\citep{Fahlman:1990:CLA:109230.107380},
\textit{weight sharing}~\citep{Nowlan:1992:SNN:148167.148169},
\textit{pruning}~\citep{Hassibi:1992:SOD:645753.668069} and
\textit{noise injection}~\citep{Holmstrom92}. In contrast to this kind
of techniques, early stopping~\citep{Morgan:1990:GPE:109230.109302}
does not aim to directly eliminate overfitting, but rather to stop
training when there are reasons to suspect it is occurring. While on
paper one might think that this is not the best way to deal with the
problem,~\cite{Bishop95} proves that it leads to similar solutions to
regularization and, in fact, it can be viewed as regularization over
time~\citep{Sjoberg95}. Moreover, its potential as a complement to
classic regularization routines and its conceptual simplicity implies
lower computational complexity and greater
effectiveness~\citep{Finnoff1993771,Raskutti:2014:ESN:2627435.2627446},
making this a particularly interesting strategy from a practical point
of view. This, the design of early stopping conditions, sets the
context within which we approach the study of regularization in {\sc
  dl}-based learning environments. With that in mind, we first review
in Section~\ref{section-related-work} the methodologies that served as
inspiration, as well as our contributions. Next,
Section~\ref{section-formal-framework} reviews the formal basis
supporting the proposal described in full in
Section~\ref{section-abstract-model}. In
Section~\ref{section-testing-frame}, we introduce the testing frame,
including both the monitoring structure and a quality metric, for the
experiments described and analyzed in
Section~\ref{section-experiments}. Finally,
Section~\ref{section-conclusions} presents our conclusions.

\section{Related work and contribution}
\label{section-related-work}

Early stopping strategies for {\sc nn}{\footnotesize s} can be roughly
classified according to whether or not a validation set, different
from the training set, is considered to monitor learning performance
using some kind of quality metric. However, because this latter is
mostly based on cross-validation, as is the case for the
\textit{mean squared error} distance, differentiation is often done 
by the use or non-use of such criteria.

\subsection{Early stopping without cross-validation}

While the simplest approach is to set the number of epochs used to generate
the model, this is not, strictly speaking, a stopping condition
because it is unrelated to the level of overfitting in learning. In
fact, some authors~\citep{Lodwich:2009:ERP:1704175.1704224} prefer the
term \textit{primary rule}, in the sense that the condition is met
\textit{a fortiori} at some point, thus providing a frame to evaluate
other \textit{secondary  rules} that cannot be guaranteed to
trigger. More formally,~\cite{Cataltepe:1999:NFL:313776.313813} show
that, in the absence of information other than the training examples,
the generalization error is an increasing function of the training
one. Thus, an optimal choice for an early stopping solution should
probably be any one associated with the minimum in this latter,
although no method has been described to compute it. For their
part,~\cite{Liu08} define a \textit{signal-to-noise-ratio figure}
({\sc snrf}) to measure the goodness-of-fit using the training error,
which enables the detection of overfitting without the use of a
separate validation set. However, there are two problems with this
criterion~\citep{Piotrowski2013} that make its application
problematic. First of all, it depends only on sample size, and so
fails to take into account the nature and quality of real-world
data. Secondly, it may produce negative values for multidimensional
{\sc nn}{\footnotesize s}, thus resulting in immediate stopping.

Other works address the question from a statistical perspective,
seeking a trade-off between {\sc nn} complexity and training
error. For linear systems,~\cite{WangVenJudd94} analyze the average
optimal stopping time from the probability density error in training
data. Unfortunately, this approach is only useful for {\sc nn}{\footnotesize s}
where output weights are being trained and there is no hidden
layer training. From a more operational perspective, substantial
research focuses on boosting {\sc ml}, which produces accurate
prediction criteria by minimizing a loss function that combines rough
and moderately inaccurate rules-of-thumb. For early stopping,
~\cite{Buhlmann03} prove optimality in the case of fixed design
regression, although with a rule that is not computable from the
data. More generally,~\cite{zhang2005} establish universal consistency
and convergence upper bounds for convex loss regularizers.

To facilitate practically useful outcomes, subsequent works simplify
the hypotheses for overfitting estimators, assuming that, instead of
convex or linear combinations of functions, the underlying loss
function belong to a \textit{reproducing kernel Hilbert
  space}\footnote{The family characterized by polynomial decreasing
  rates of step sizes.} ({\sc rkhs}). So,~\cite{Yao2007} report faster
and optimal convergence rates, but do not analyze lower bounds. More
recent works overcome this drawback when the upper bounds are improved
by studying the eigenvalues of the kernel matrix, thus yielding
minimax optimal rates of estimation for a broad class of loss
functions and various kernel functions. This is, for example, the case
of conjugate gradient~\citep{Blanchard10},
boosting~\citep{Raskutti:2014:ESN:2627435.2627446} or gradient
descent~\citep{Blanchard2018} algorithms. Unfortunately, such proposals
require an unnaturally large --asymptotic-- sample size, which not only
comes with a considerable computational overhead, but also tends to be
suboptimal in practice.~\cite{Wei17}, on the other hand, provide
similar guarantees without having to calculate all the eigenvalues,
although the procedure is unimplementable in practice.

\subsection{Early stopping with cross-validation}

The idea here is to use the accuracy reached by a model on the
validation dataset as an indicator of its generalization
error~\citep{Morgan:1990:GPE:109230.109302}. As the goal is to
minimize the error, learning is stopped when deteriorating performance
is interpreted as a sign of overfitting. This poses a non-obvious
question, as the error in a validation set often has more than one
local minimum, so the problem is solvable only when the initial
weights of the {\sc nn} are small
enough~\citep{Baldi:1991:TEG:1350931.1350941}. Otherwise, and by
fixing training collection and initial weights,~\cite{Dodier96} proves
that for finite validation sets there is a dispersion of stopping
points around the best one, i.e., the most probable with the least
generalization error, which increases the expected generalization
error. Then,~\cite{Amari97} statistically estimate the optimal split
proportion between validation and training data to obtain optimum
performance, although the proposal holds asymptotically and is not
practical. Such a lack of simple formal solutions to address the
halting issue has supported the implementation of a wide variety of
\textit{na\"ive} strategies, all of which are largely
unreliable. Typically they involve stopping the first time a given
accuracy value is attained or after a number of iterations is
performed without measurable improvement~\citep{Shao11}. It may also
be plausible to stop when performance begins to become erratic or the
curves that reflect its evolution in training and validation sets
begin to cross~\citep{Lodwich:2009:ERP:1704175.1704224}.

It is within this frame that~\cite{Prechelt97automaticearly}
empirically reports three families of stopping criteria, indexed by a
tuning parameter ($\alpha$) to permit small excursions in the
validation error: \textit{generalization loss} ($\mbox{\sc
  gl}_\alpha$), \textit{productivity quotient} ($\mbox{\sc
  pq}_\alpha$) and \textit{uninterrupted progress} ($\mbox{\sc
  up}_\alpha$). The first of these criteria stops learning as soon as
the relative increase in the validation error exceeds $\alpha$. To
avoid stopping when the training error drops rapidly, because this
would give the generalization losses a chance to be repaired, the
second condition waits until the ratio of {\sc gl} to training
progress in a strip of epochs is greater than $\alpha$. The third
criterion, {\sc up}, suggests stopping when the validation error keeps
increasing for $\alpha$ successive strips.

Although Prechelt's metrics should allow us to make a choice based on
efficiency, effectiveness or robustness concerns, no metric seems to
predominate in terms of average generalization performance. So, while
longer training time appears to improve generalization performance in
all cases, the {\sc pq} proves to be most cost-effective only for
sufficiently small {\sc nn}{\footnotesize s}. Either way, Prechelt's
proposal has inspired a good part of further work in practical early
stopping in {\sc dl}, and also in \textit{evolutionary algorithms},
typically through \textit{genetic programming} ({\sc gp}) techniques
in which the counterpart of the \textit{loss function} is identified
with the \textit{fitness
  function}. Specifically,~\cite{Foreman:2005:POG:1068009.1068307}
study the use of online indicators from the Prechelt's metrics with
cross-validation as \textit{canary function}, and as also done in a
later work by~\cite{Vanneschi:2010:MBO:1830483.1830643}, they raise
the interest of exploring the \textit{correlation} between fitness in
the training set and the validation one. Alas, to the latter
possibility,~\citep{Vanneschi:2010:MBO:1830483.1830643} merely apply
the concept from a visual point of view, with no formal basis. Turning
to~\cite{Foreman:2005:POG:1068009.1068307}, they limit themselves to a
brief discussion without including technical details, which concludes
by proposing the combination of a correlation coefficient --which they
do not even identify-- as the primary condition and {\sc pq} as the
complementary one. \cite{Tuite:2011:ESC:2001858.2001971}, however,
disagree with this approach and report the {\sc pq} to be less
effective in {\sc gp}, a behavior
that~\cite{Nguyen:2012:WSI:2437054.2437098} associate with the use of
excessively small values for the $\alpha$ tuning parameter. This lack
of consensus is also reflected in the {\sc nn} sphere,
where~\cite{Piotrowski2013} adopt the simplest {\sc gl} and terminate
training when the validation error exceeds its previously noted
minimum by 20\%. More recently,~\cite{WangYan2017} have extended
Prechelt's approach by combining the probability density function
error of the unlabeled sample data and the validation error of the
labeled ones, although the impact on the reduction of the
generalization error seems limited.

A different view of the issue is to seek to improve efficiency by
selecting the validation series on the basis of its mean dynamic
correlation with forecast performances, in the complete training
data~\citep{Michalak:2005}. However, the validation set needs to be
sufficiently large, even when the observed enhancement is
statistically significant. Nonetheless, those authors implicitly
suggest the grouping of single conditions to generate a compound rule
that stops when one of those so determines, an approach later explored
by~\cite{Lodwich:2009:ERP:1704175.1704224} for six families of single
criteria, including Prechelt's ones. Their tests confirm the apparent
superiority of {\sc pq} over the rest of the individual rules, while
the top results seem to require criteria to be coupled, with ensembles
based on $\mbox{\sc pq}_3$ performing best.

\subsection{Our contribution}

Against this backdrop, we formally describe and test an early stopping
procedure --which we have called {\sc coi}, for {\sc c}orrelation of
{\sc o}nline {\sc i}ndicators-- with a view to reducing the risk of a
misguided early stopping, but also to improving the trade-off between
training effort and generalization error. In order to provide the
adaptability to each learning process that this requires, our proposal
exploits the properties of potentially complementary halting rules,
with the sole condition that the latter are organized as a repository
of indicators for some canary function regarding overfitting
phenomena. The technique takes a correlation coefficient as a measure
of the degree of agreement in the diagnosis by the individual rules,
which are assumed to be defined by sets of variables that are mutually
independent. This allows us to put ourselves in a context in which the
\textit{principle of the common cause} holds, thus supporting the
correctness of a stopping criterion based on sufficiently broad
agreement between at least two of the diagnoses. Thus, once the user
has set a correlation coefficient and a confidence threshold for it,
the halting decision is made over time according to the highest
alignment between a pair of individual indicators.

We choose the {\em natural language processing} ({\sc nlp}), a set of
computational methods aimed at processing and understanding human
languages, as test environment. This domain, increasingly linked to
data-driven models~\citep{Jones1994}, is featured by a growing
interest in using {\sc dl}
techniques~\citep{Goldberg:2017,Young18}. The main reason is the
ability to automatically generate multilevel features to produce
distributed representations from texts at both word~\citep{Mikolov10}
and syntactic~\citep{Collobert:2011} levels. Also known as
\textit{embeddings}, these features can be used as a first processing
layer in {\sc dl}-based {\sc nlp} models\footnote{While most of the
  recent work in {\sc nn}{\footnotesize s} applied to {\sc nlp} relies
  on end-to-end training without resorting to extra linguistic
  resources~\citep{Manning:2015:CLD:2893320.2893326}, evidence
  suggests that integrating symbolic knowledge into models aids
  learning~\citep{Faruqui-2015,Plank-Klerke-2019-lexical}.}. This
brings a revolution because, apart from minimizing the need for
time-intensive engineering work and hand-designed resources, the
latter are often incomplete or sparse. It is therefore expected that
advances in computational power and
parallelization~\citep{Coates:2013,Raina:2009}, together with the
availability of large training datasets, offer a bright future for
these methods, which justifies our choice.

\section{Formal framework}
\label{section-formal-framework}

Below we review the fundamentals of the abstract model that will be
introduced later, namely the use of online indicators for canary
functions as an early warning mechanism for assessing the risk of
overfitting in {\sc nn}{\footnotesize s}, and the definition of a
trusted decision-making scenario from the correlation of different
criteria.

\subsection{Canary functions and online indicators for overfitting identification}

To recognize when overfitting may be starting to occur, precise
characterization is required to capture the operational meaning behind
this idea. A close reference is the notion of \textit{canary
  function}, introduced to solve the issue in {\sc
  gp}~\citep{evett:1998:GPsqp}. Such functions differ from the fitness
ones in that their value begins to degrade at about the same time
overfitting starts. Following this
path,~\cite{Foreman:2005:POG:1068009.1068307} take advantage of
cross-validation approaches, using performance divergence between
training and validation sets to determine when overfitting occurs.
Unfortunately, a problem arises with practical implementations, since
it is difficult to estimate the actual scale of such deviation from
the local perspective of an iterative learning scheme. Drawing
inspiration from the stopping conditions described
by~\cite{Prechelt97automaticearly} for {\sc nn}{\footnotesize s}, the
authors use Boolean functions --called \textit{online indicators}--
to evaluate from data available at each generation of a {\sc gp} run
whether that disparity is significant. In addition, they hypothesize
that the correlation coefficient between the fitness of the best
individual of the generation with respect to the training set, and the
fitness of the best individual of the generation with respect to the
validation one at the end of each generation should also be a strong
divergence signal.

Following this path, we adapt the concept of a canary function to {\sc
  nn}{\footnotesize s} by using a loss function instead of a fitness
one, as that permits us to reflect performance discrepancy between
the training and validation sets. The notion of online indicator is
then readjusted accordingly, allowing Prechelt's stopping criteria to
be applied to the task, as well as any correlation coefficient for
training and validation set performance.

\subsection{An explanatory basis for correlating criteria}

The core of our work rests on the idea that the stopping decisions for
several indicators using the same canary function can yield
complementary insights and thus increasing the reliability of
individual criteria, i.e. predictive power can be enhanced through an
appropriate combination of such criteria. It is therefore
necessary to determine the conditions under which a solution of this
type is applicable. This means, first and foremost, establishing a
theoretical framework for their analysis, on which to then define a
well-founded reasoning scheme to ensure the correctness of our
approach.

\subsubsection{The principle of the common cause}

In essence, our initial argument relies on a metaphysical claim that
refers to similar state-of-the-art interpretations in the philosophy
of causation~\citep{Mill1868,Russell1948}, stating that:

\begin{center}
  {\textit{``if an improbable coincidence has occurred, there
      must exist a common cause''}}
\end{center}  

\noindent However, since this says nothing about how to characterize
cause/effect relationships, it is not helpful for implementation
purposes, although it does give our proposal meaning.~\cite{Reichenbach1956}
fills this gap by a methodological claim based on a 
probabilistic criterion. Suppose events $A$ and $B$ are positively
correlated, namely
\begin{equation}
\label{eq-CPP-correlation}
  p(A \; \& \; B) > p(A) \; p(B)
\end{equation}  
\noindent and assume that neither $A$ nor $B$ is a cause of the
other. Then Reichenbach maintains that there is a common cause $C$, of
$A$ and $B$, meeting the conditions:
\begin{equation}
\label{eq-CPP-screening-off-1}  
  p(A \; \& \; B \mid C) = p(A \mid C) \; p(B \mid C) 
\end{equation}
\begin{equation}
\label{eq-CPP-screening-off-2}    
    p(A \; \& \; B \mid \neg C) = p(A \mid \neg C) \; p(B \mid \neg C)
\end{equation}  
\begin{equation}
\label{eq-CPP-C-cause-of-A}    
  p(A \mid C) > p(A \mid \neg C)
\end{equation}    
\begin{equation}
\label{eq-CPP-C-cause-of-B}    
    p(B \mid C) > p(B \mid \neg C)
\end{equation}
\noindent In particular, Eq.~\ref{eq-CPP-screening-off-1}
(resp. Eq.~\ref{eq-CPP-screening-off-2}) is construed to mean that $C$
(resp. $\neg C$) \textit{screens off} the correlation between $A$ and
$B$. This correlation disappears when we take into account the
common cause, which is to say that $A$ and $B$ are then
probabilistically independent. Meanwhile, 
Eq.~\ref{eq-CPP-C-cause-of-A} (resp. Eq.~\ref{eq-CPP-C-cause-of-B}) states
that $C$ is a cause for $A$ (resp. $B$). Altogether,
Eqs.~\ref{eq-CPP-screening-off-1}-\ref{eq-CPP-C-cause-of-B} characterize what is defined as a \textit{conjunctive
  fork} $ACB$, namely a causal structure
$A \leftarrow C \rightarrow B$.

According to the
state-of-the-art~\citep{Reichenbach1956,SanPedro2009}, we will refer
to the metaphysical claim as the \textit{postulate of the common
  cause} ({\sc p}os{\sc cc}), to the methodological claim as the
\textit{criterion for common causes} ({\sc c}rit{\sc cc}) and to the
conjunction of both claims as the \textit{principle of the common
  cause} ({\sc pcc}). This provides us with a theoretical framework
for causation analysis by introducing \textit{probabilistic
  causality}, a notion in philosophy whose main message is that causes
change the probabilities of their
effects~\citep{Cartwright1979,Eells1991,Skyrms1981,Suppes1970}.

\subsubsection{The explanatory power of different causal assumptions}

Unfortunately, although Reichenbach's view confers the predictibility
necessary to give support to our proposal, the {\sc pcc} cannot be
accepted at face value because the {\sc c}rit{\sc cc} may fail. This
may occur particularly when the correlated events are logically
related or spatio-temporally overlapping~\citep{Hitchcock2012}, and
also when there is an attempt to apply it in domains where classic
causal intuition is no longer valid. The latter scenario is typically
the case of quantum mechanics, whose recent success has renewed
interest in probabilistic causation to the detriment of
\textit{deterministic} approaches~\citep{Hume1904} --also called
\textit{regular} approaches-- according to which causes are
invariably followed by their effects.

It thus becomes evident that {\sc c}rit{\sc cc} is not a sufficient
condition for a conjunctive fork\footnote{The relations expressed by
  Eqs.~\ref{eq-CPP-C-cause-of-A} and~\ref{eq-CPP-C-cause-of-B} are
  symmetrical. So, for example, Eq.~\ref{eq-CPP-C-cause-of-A} could be
  rewritten equivalently as $ p(C \mid A) > p(C \mid \neg A)$, in
  which case {\sc c}rit{\sc cc} also defines the conditions that apply
  to the linear causal diagram $A \rightarrow C \rightarrow B$.},
which justifies the formulation of theories close to {\sc pcc} but
requiring that the likely cause occurs at an earlier time than the
correlated
effects~\citep{Arntzenius93,Frassen82b}. Alternatively,~\cite{Hofer1998,Hofer2013}
seek a screening-off common cause, as described by Reichenbach, by
extending the original probability space to another causally closed
space, while the question remains whether the common causes thus
detected represent anything physically real. Finally, other
authors~\citep{Cartwright1988,salmon1984scientific} simply accept the
validity of {\sc pcc}, but only under a revision of {\sc c}rit{\sc
  cc}. In short, there is no agreement regarding the status of {\sc
  pcc} and its relationship to the notion of common
cause~\citep{Hofer2013,Forster14}.

Thankfully, in deterministic contexts with mutually independent events
caused by an earlier one, the {\sc pcc} holds even when the
operational conditions do not allow for trivialization. It can
therefore be interpreted to reflect the imperfect knowledge of the
causal system, thus justifying its application~\citep{Arntzenius10}.

\section{Abstract model}
\label{section-abstract-model}

The theoretical foundations of our proposal are explained below for
later interpretation from an operational point of view. We first
explain the working hypotheses and notational support, and then
formally describe our abstract model to finally establish its
correctness.

\subsection{Working hypotheses and notational support}

Let $\mathbb{R}$ be the real number line and $\mathbb{N}$ the natural
number line, with $0 \not\in \mathbb{N}$. We are given a fixed
sampling database ${\mathcal D} = {\mathcal D}_{\textit{tr}} \cup
{\mathcal D}_{\textit{va}} \cup {\mathcal D}_{\textit{te}}$, referring
respectively to test, training and validation sets. Each ${\mathcal
  D}_{j} := \{(x^i_j,f^i_j)\}_{i=1}^{N_j}$ includes inputs $x^i_j \in
\mathbb{R}^L$, outputs $f^i_j \in \mathbb{R}^M$ and $L, M, N_j \in
\mathbb{N}$, with $j \in \{\textit{tr}, {\textit{va}},
       {\textit{te}}\}$. The outputs are assumed to be generated from
       the inputs in a deterministic causal system according to some
       unknown distribution $f$ and, hence, $f^i_j = f(x^i_j), \forall
       i \in [1,N_j]$. Depending on the {\sc ml} strategy used,
       ${\mathcal D_{\textit{te}}}$ may or may not be empty.

A {\sc dl}-based learning scheme, hereafter a \textit{kernel},
is denoted by ${\mathcal K}={\mathcal N}^{\pi}[{\mathcal D}]$, with
${\mathcal N}$ a {\sc nn}, ${\mathcal D} = {\mathcal D}_{\textit{tr}}
\cup {\mathcal D}_{\textit{va}} \cup {\mathcal D}_{\textit{te}}$ a
sampling database and $\pi$ a collection of parameters that includes a
\textit{propagation function}, which can add a bias term to its result
but also possibly an \textit{activation} function. We then use
${\mathcal K}[e]$ to refer to the \textit{model} generated at epoch $e$
from such a learning scheme, and $E_j^{\mathcal K}[e]$ to refer to the
value for it of a \textit{loss function} --also referred to as an
\textit{error function}-- $E$ in ${\mathcal D}_{j} \subseteq
       {\mathcal D}, j \in \{\textit{tr}, {\textit{va}},
       {\textit{te}}\}$.

Having set $\Upsilon_c := \{c_i\}_{i \in I}$ a finite family of online
indicators on a canary function $c$ and ${\mathcal K} = {\mathcal
  N}^{\pi}[{\mathcal D}]$ a kernel, we then refer to the value of $c_i
\in \Upsilon_c$ at epoch $e$ as $c_i^{\mathcal K}[e]$, and the epoch
at which $c_i$ verifies and the training of ${\mathcal K}$ stops as
$s_{c_i}^{\mathcal K}$. The model generated at this latter is labelled
as a \textit{run} ${\mathcal R} = {\mathcal N}^{\pi}_{c_i}[{\mathcal
    D}]$, namely ${\mathcal R} := {\mathcal K}[s_{c_i}^{\mathcal K}]$.
More intuitively, a run is the result of a specific learning process
from a kernel ${\mathcal K}$, the completion of which is determined by
an online indicator $c_i$ defined on a given canary function $c$. We
can then naturally introduce $E_j^{\mathcal R}$ as $E_j^{\mathcal
  K}[s_{c_i}^{\mathcal K}]$, i.e. the value of the error function
measured for the model associated with the run ${\mathcal R}$ when it
applies on a database ${\mathcal D}_{j} \subseteq {\mathcal D}, j \in
\{\textit{tr}, {\textit{va}}, {\textit{te}}\}$.

Within the context described above, the precision of an indicator $c_i
\in \Upsilon_c$ in relation to the training procedure represented by
the kernel ${\mathcal K}$ will be all the greater the closer the epoch
at which its termination occurs ($s_{c_i}^{\mathcal K}$) to the one
that determines the time at which the overfitting actually arises, the
latter denoted by $\mbox{\em \oe}^{\mathcal K}$. For testing purposes,
the exact location of the epoch $\mbox{\em \oe}^{\mathcal K}$ can be
estimated from a sufficiently large number of epochs in ${\mathcal
  K}$, referred to as its \textit{horizon}, by selecting the iteration
that is associated with the lowest validation error. 


\subsection{Our proposal}

Given a kernel ${\mathcal K}={\mathcal N}^{\pi}[{\mathcal D}]$, namely
a learning process on a sampling database ${\mathcal D}$ using a
learner ${\mathcal N}$ according to a setting $\pi$, we seek to
identify the epoch $\mbox{\em \oe}^{\mathcal K}$ in which training
should stop to prevent overfitting. Taking the values provided by a
canary function $c$ as basis for decision making on the diagnosis of
this type of phenomena, and $\Upsilon_c := \{c_i\}_{i \in I}$ as a
finite set of associated and pairwise independent online indicators
defined on it, the idea is to provide robustness for early stopping by
exploiting their consensus.  To this end, we resort to a correlation
coefficient $\aleph$, which we assume to be compatible with the
underlying distribution of the values generated by such indicators.

Thus, training stops when at least two such indicators so advise
within a \textit{training strip of length $k$}, i.e., a sequence of
$k$ epochs numbered $n + 1 \cdots n + k$ where $n$ is divisible by
$k$, in which their $\aleph$-correlation is above a threshold. In
particular, the indicators do not need to agree on the same epochs,
thereby introducing a limited level of asynchronicity in order to make
the condition more flexible. Against this backdrop, we first capture
that notion of flexible $\aleph$-correlation within a training strip
of length $k$ at epoch $e$ by:

\begin{equation}
\label{eq-COIk-criterion}  
c_{\textit{coi}[\aleph,k]}^{\mathcal K}[e] :=
\max_{\iota, \jmath \in I} \{0, \; \aleph_{{\mathcal C}_{\iota}^{\mathcal K}[k,e]}^{{\mathcal C}_{\jmath}^{\mathcal K}[k,e]} \; \mbox{\em such that} \; \exists \; i, j \in [k-1, 0] \; \mbox{\em for which} \; c_{\imath}^{\mathcal K}[e-i] \; \mbox{\em and} \; c_{\jmath}^{\mathcal K}[e-j] \}
\end{equation}
\noindent with ${\mathcal C}_{\imath}^{\mathcal K}[k,e] :=
\{c_{\imath}^{\mathcal K}[e-k+1], \dots, c_{\imath}^{\mathcal K}[e]\},
\; \forall \imath \in I$. Having fixed a minimum value
$\alpha_\aleph$, the corresponding online indicator is then given by
the Boolean expression:

\begin{equation}
  \label{eq-COIk-online-indicator}
  c_{\textit{coi}[\aleph,k,\alpha_\aleph]}^{\mathcal K}[e] := [c_{\textit{coi}[\aleph,k]}^{\mathcal K}[e] > \alpha_\aleph]
\end{equation}

\noindent which is true when the value
$c_{\textit{coi}[\aleph,k]}^{\mathcal K}[e]$ calculated from
Eq.~\ref{eq-COIk-criterion} is greater than $\alpha_\aleph$, and false otherwise.

\subsection{Correctness}

Once our working hypotheses have been established, the goal is to
prove that the positive correlation between online indicators
increases the reliability of the diagnosis of overfitting
phenomena. Firstly, this implies verifying the conditions of
applicability for the {\sc c}rit{\sc cc} --the methodological claim
of the {\sc pcc}--, which will allow us to conclude that there is a
common cause behind such a statistical relationship and that this is not the
result of a dependency link between the indicators. Afterwards, it will
be sufficient to identify that such common cause is the degradation of
values provided by the canary function considered, which we assume to
be at the origin of a generalization loss and, therefore, also of a
potential overfitting of the {\sc ml} process.

\subsubsection{Applicability for the {\sc c}rit{\sc cc}} 

We are talking in this case about an immediate conclusion because our causal
structure is supposedly regular and the observed variables, namely the
online indicators, are also assumed to be free from causal
relationships between them. In other words, our working hypotheses
ensure that the {\sc c}rit{\sc cc} is interpretable as a necessary and
sufficient condition on common
causes~\citep{Arntzenius10,SanPedro2009}.

\subsubsection{Identification of the common cause}

In the previously set correctness context, the only remaining task is
to identify the common cause behind any positive correlation as the
increase in the canary function of the performance discrepancy between
the training sets and the validation ones. We are again talking about
an immediate conclusion, this time derived from the very notion of an
online indicator, which is essentially a Boolean that determines
whether performance reduction is significant enough to suspect that
overfitting occurs. In other words, the common cause is the observed
decline in learning acquisition. \\

The {\sc pcc} thus provides, by means of a simple probabilistic
approach to causation, a way of enhancing reliability in the diagnosis
of overfitting phenomena independent of temporary variations in the error
function over the course of an {\sc ml} process. Put another way,
decision making based on the correlation of complementary stopping
criteria diminishes the risk of wrong appraisal, often resulting from
simple local irregularities in the values generated by the canary
functions.

\section{Testing frame}
\label{section-testing-frame}

To support our conclusions, we design a categorizing protocol for
early stopping criteria based on a quality metric. In order to
guarantee its reliability, the conditions under which the experiments
take place are normalized by introducing a specific monitoring
architecture for data collection.

\subsection{Monitoring structure}

With the purpose of enabling the monitoring of the trials in our
testing frame, it is first necessary to design a structure that allows
the online indicators to compete on fair terms. This implies
organizing our study around a family of local testing frames that
encapsulate those common assessment conditions according to different
combinations of learners and sampling databases, i.e. different
kernels. Against this backdrop, our evaluation basis is the previously
entered run structure ${\mathcal N}_{c_i}^{\pi}[{\mathcal D}]$,
essentially a model generated from a kernel ${\mathcal
  N}^{\pi}[{\mathcal D}]$, namely a training task on a sampling
database ${\mathcal D}$ according to a given learning architecture
${\mathcal N}$ for a particular setting $\pi$, and a stopping
criterion fixed by an online indicator $c_i$ within a collection
$\Upsilon_c$ in which all have been defined on the same canary function
$c$.

Obviously, we also need to set a reference base for our research findings
which should be shared by the set of local evaluation settings
considered in order to provide a comprehensible general overview of our
proposal. We then assume $c_{\wp[h]}$ to be an optimal online
indicator for a canary function $c$, with values provided by an
omniscient oracle on a horizon $h$ in such a way that, whatever the
run ${\mathcal N}_{c_{\wp[h]}}^{\pi}[{\mathcal D}]$
considered, $\mbox{\em \oe}^{\mathcal K} = s_{c_{\wp[h]}}^{\mathcal K}$
for its corresponding kernel ${\mathcal K}={\mathcal N}^{\pi}[{\mathcal
    D}]$. In other words, the epoch $\mbox{\em \oe}^{\mathcal K}$ in
which the overfitting occurs, is the same epoch
$s_{c_{\wp[h]}}^{\mathcal K}$ at which the online indicator
$c_{\wp[h]}$ indicates it, thereby justifying its referential
character.

Once a family $I$ of kernels is fixed, the study of a 
learning scheme $i \in I$ is formalized around the notion of
\textit{local testing frame} $\mathcal{L}_i^{\Upsilon_c}:= \{{\mathcal
  N}^{\pi}_{c_{\iota}}[{\mathcal D}], \; c_\iota \in \Upsilon_c \cup
\{c_{\wp[h]}\}\}$. We are then talking about a set of runs defined on
the same kernel ${\mathcal N}^{\pi}[{\mathcal D}]$ and only
distinguishable by their online indicator $c_{\iota}$ taken
from a finite set $\Upsilon_c$ related to a canary function $c$, which
also includes ${\mathcal N}_{c_{\wp[h]}}^{\pi}[{\mathcal D}]$ as 
\textit{baseline}. That way, the collection ${\mathcal L} := \{
\mathcal{L}_{i}^{\Upsilon_c}\}_{i \in I}$ of local testing frames
becomes naturally our monitoring structure.

\subsection{Performance metric}

According to the principle of \textit{maximum expected utility} ({\sc
  meu})~\citep{Meek:2002:LSM:944790.944798}, we interpret the performance
associated with a run as the search for a satisfactory cost/benefit
trade-off, involving both quantitative and qualitative
considerations. For the quantitative considerations, we take into
account the final computational cost, which depends on the degree of
control exercised by the user over the {\sc ml} process. In its
absence, as in our case, those costs are the sum of the data acquisition,
error and model induction costs~\citep{WeissTian08}.  Since the kernel
${\mathcal N}^{\pi}[{\mathcal D}]$ is common to all runs in a local
testing frame $\mathcal{L}^{\Upsilon_c}:= \{{\mathcal
  N}^{\pi}_{c_{\iota}}[{\mathcal D}], \; c_\iota \in \Upsilon_c \cup
\{c_{\wp[h]}\}\}$, so too are their acquisition costs, while error
(resp. model induction) costs are proportional to the cumulative error
for each run in the validation dataset (resp. the number of epochs
applied). As for the qualitative considerations, baseline runs provide
an easy to understand gold standard within local testing frames, thereby
offering a simple and practical framework to assess online indicators.

\begin{df}
\label{def-normalized-MEU-signed-deviation}
Let
${\mathcal L}^{\Upsilon_c}:= \{{\mathcal
  N}^{\pi}_{c_{\iota}}[{\mathcal D}], \; c_\iota \in \Upsilon_c \cup
\{c_{\wp[h]}\}\}$ be a local testing frame on a kernel
${\mathcal K} = {\mathcal N}^{\pi}[{\mathcal D}]$, with baseline
${\mathcal B}$. We define the {\em normalized {\sc meu} signed
  deviation} {\em (}\nemesid{}{\em )} of a run
$ {\mathcal R} := {\mathcal N}^{\pi}_{c_{\iota}}[{\mathcal D}] \in
{\mathcal L}^{\Upsilon_c}$ as:
\begin{equation}
\label{eq-normalized-MEU-signed-deviation}
\Phi^{\omega_{mi}}_{\omega_{\textit{ea}}}({\mathcal R},{\mathcal L}^{\Upsilon_c}) :=
\left\{ \begin{array}{ll}
\frac{\mbox{\sc c}^{\omega_{\textit{mi}}}_{\omega_{\textit{ea}}}({\mathcal R}) - \mbox{\sc c}^{\omega_{\textit{mi}}}_{\omega_{\textit{ea}}}({\mathcal B}) + \mbox{\sc mcdb}^{\omega_{\textit{mi}}}_{\omega_{\textit{ea}}}({\mathcal L}^{\Upsilon_c})} 
           {\mbox{\sc mcdb}^{\omega_{mi}}_{\omega_{\textit{ea}}}({\mathcal L}^{\Upsilon_c})} -1 & \mbox{if } \mbox{\sc mcdb}^{\omega_{\textit{mi}}}_{\omega_{\textit{ea}}}({\mathcal L}^{\Upsilon_c}) \neq 0 \\
0 & \mbox{otherwise}           
         \end{array}
         \right.
\end{equation}
\noindent where
\begin{equation}
\label{eq-maximum-cost-deviation-from-baseline}
\mbox{\sc mcdb}^{\omega_{\textit{mi}}}_{\omega_{\textit{ea}}}({\mathcal L}^{\Upsilon_c}) := \max_{{\mathcal R} \in {\mathcal L}^{\Upsilon_c}} \abs{\mbox{\sc c}^{\omega_{\textit{mi}}}_{\omega_{\textit{ea}}}({\mathcal R}) - \mbox{\sc c}^{\omega_{\textit{mi}}}_{\omega_{\textit{ea}}}({\mathcal B})} 
\end{equation}
\noindent captures the {\em maximum cost deviation from the baseline} in
${\mathcal L}^{\Upsilon_c}$ and
\begin{equation}
\label{eq-cost}  
\mbox{\sc c}^{\omega_{\textit{mi}}}_{\omega_{\textit{ea}}}({\mathcal R}):= \omega_{\textit{mi}} \ast s_{c_{\iota}}^{\mathcal K} + 
\omega_{\textit{ea}} \ast E^{\mathcal R}_{\textit{va}}, \mbox{ such that }
\omega_{\textit{mi}}, \omega_{\textit{ea}} \in (0, 1]
\end{equation}
\noindent is the {\em cost} of
${\mathcal R} \in {\mathcal L}^{\Upsilon_c}$, expressed as the sum of
the model induction and error acquisition costs, with processing
weights for each epoch and error assumed to be $\omega_{\textit{mi}}$ and
$\omega_{\textit{ea}}$, respectively.
\end{df}

As the name implies, \nemesid{} quantifies and normalizes the signed
deviation of a run from its baseline in {\sc meu} terms. Since such
baselines are always built on the online indicator supported by the
omniscient oracle, this gives us an effective means of estimating the
performance of any other indicator through different local testing
frames, something that an absolute measure cannot do.

More formally, and once a local testing frame is fixed, \nemesid{} is
strictly increasing with respect to the cost of the runs. Since the
minimum (resp. maximum) is reached in runs with the lowest
(resp. highest) cost compared with the baseline, the codomain is the
interval $[-1, 1]$. In contrast, the modulus of this metric is
strictly decreasing with respect to the utility of the runs, and
because the value is null for the baseline, the codomain is now the
interval $[0,1]$. Accordingly, the lower its real and absolute
\nemesid{} values, the better a run performs. In other words, the most
efficient online indicators are those providing rates as close to zero
as possible, preferably negative.

\section{Experimental results}
\label{section-experiments}

Given a kernel ${\mathcal N}^{\pi}[{\mathcal D}]$ representing an {\sc
  ml} process over a series of epochs, and taking cross-validation as the
canary function $c$, the goal is to illustrate how well our online
indicator performs in detecting overfitting. 

\subsection{Case study}

We focus on parser generation in the domain of {\sc nlp}, not only
because of the complexity of annotation in building training data and
then learning to capture the relationships at training time, but also
because parsers serve as input for other {\sc nlp}
functionalities~\citep{Jurafsky2009}. From a marked-up text provided
by a \textit{part-of-speech} ({\sc pos})\footnote{A {\sc pos} is a
  category of words which have similar grammatical properties. Words
  that are assigned to the same {\sc pos} generally display similar
  behavior in terms of syntax, i.e., they play analogous roles within
  the grammatical structure of sentences. The same applies in terms of
  morphology, in that words undergo inflection for similar
  properties. Common English {\sc pos} labels are noun, verb,
  adjective, adverb, pronoun, preposition, conjunction, interjection,
  numeral, article or determiner.} tagger, a parser identifies which
words modify others and how, resulting in a sentence structure. The
rise of {\sc dl} technologies has propelled the interest in
dependency-based methods~\citep{Tesniere:1959}. In contrast to classic
\textit{constituency parses}, which hierarchically break text into
terminals --words-- and non-terminals --syntagms--,
\textit{dependency parses} look at the relationships between pairs of
words to produce trees of terminal nodes. The resulting
\textit{dependency arcs} hold between a \textit{head}, which
determines the behavior of the pair, and a \textit{dependent}, which
acts as its modifier or
complement~\citep{Kubler:2009:DP:1538443}. These arcs are labeled to
supply additional environmental features and are projected to
embeddings, that can be more efficiently exploited by semantic-based
{\sc nlp} applications\footnote{Here we include paraphrase
  acquisition~\citep{Shinyama:2002:APA:1289189.1289218}, knowledge
  extraction~\citep{Culotta-Sorensen-2004-dependency}, discourse
  understanding~\citep{Sagae-2009-analysis} or machine
  translation~\citep{Ding-Palmer-2005-machine}.} through {\sc
  nn}{\footnotesize s} than using other type of
techniques~\citep{Collobert:2011}. All of this fully justifies the
appropriateness of our case study.

\subsection{Required resources }

The objective is to select a set of linguistic and
software resources that guarantee a reliable and trustworthy
experimental evaluation, taking into account that our operational context
is the generation of {\sc dl}-based dependency parsers. To this end,
we focus our attention on the most representative corpora and
parser generators in the {\sc nlp} community. 

\subsubsection{Linguistic resources}

In accordance with the representativeness requirement we have just
outlined, these are built from the collection of tree-banks provided
by \textit{Universal Dependencies} ({\sc ud})~\citep{Marneffe2014}, an
international cooperative project that provides consistent annotation
of grammar, including syntactic dependencies and also {\sc pos} and
morphological features, and is freely available online
(\textit{https://universaldependencies.org/}).

The {\sc ud} treebanks are a collection of parsing datasets manually
annotated using a revised version of the {\sc c}o{\sc nll-x}
format~\citep{Buchholz2006} called {\sc c}o{\sc nll-u}
(\textit{https://universaldependencies.org/format.html}) and usually
partitioned into training, validation and test sets. Taken as a whole,
they are a comprehensive repository for numerous human languages,
which allows our experimental scope to be broadened beyond the usual
resource-rich languages --Chinese ({\sc zh}), English ({\sc en}),
Farsi ({\sc fa}), French ({\sc fr}), German ({\sc de}), Hindi ({\sc
  hi}), Japanese ({\sc ja}), Polish ({\sc pl}), Portuguese ({\sc pt}),
Russian ({\sc ru}) and Spanish ({\sc es})-- to also include
resource-poor languages --Catalan ({\sc ca}), Basque ({\sc eu}),
Galician ({\sc gl}) and Serbian ({\sc sr}). This also ensures that the
corpora selected are representative of a wide variety of language
families and subfamilies with very different characteristics,
including Indo-European (Germanic, Romance and Slavic), Japonic and
Sino-Tibetan, as well as a language isolate\footnote{A language not
  demonstrated to have descended from a common ancestor with any other
  language.} ({\sc eu}). Moreover, each language dataset covers
different knowledge domains (e.g. legal, news, poetry, wiki, etc), in
which case the corresponding training, validation and test sets are
aggregated into one set. Whenever possible, erroneous entries are
discarded and non-standard data are cleaned. All of this contributes
to the aim of defining a reliable testing space for {\sc ml}.

Tables~\ref{tabla-resource-rich-languages-sampling-data-sets}
and~\ref{tabla-resource-poor-languages-sampling-data-sets} summarize
details of the databases for resource-rich and resource-poor
languages, respectively, differentiating between partitions;
\textit{italic} characters indicate when a validation set is not
available and the test set is used instead. Details include the
average number of sentences, tokens per sentence and percentages of
\textit{unknown tokens} and \textit{unique tokens}\footnote{Unknown tokens are
tokens included in the validation set but not in the training one, and
\textit{unique tokens} are calculated as the quotient between the
number of different tokens and the total number of tokens.}.

\begin{table}[]
\begin{center}
\begin{tabular}{lllrrrrrrr}
\hline
                                                                                  &                      & & \multicolumn{1}{c}{{\bf\scshape de}} & \multicolumn{1}{c}{{\bf\scshape en}} & \multicolumn{1}{c}{{\bf\scshape es}} & \multicolumn{1}{c}{{\bf\scshape fr}} & \multicolumn{1}{c}{{\bf\scshape pl}} & \multicolumn{1}{c}{{\bf\scshape pt}} & \multicolumn{1}{c}{{\bf\scshape ru}} \\ \hline
\multirow{3}{*}{\rotatebox[origin=c]{90}{\scalebox{.6}{\sc \textbf{training}}}}   & sentences  &    & 166,849                               & 21,253                                & 28,492                                & 18,640                                & 31,496                                & 17,992                                & 54,099                                \\
                                                                                  & unique tokens & (\%)  & 6.49                               & 7.89                               & 7.92                               & 10.30                              & 17.74                              & 9.55                               & 12.69                              \\
                                                                                  & sentence size & & 18.08                                & 17.65                                & 29.03                                & 23.85                                & 12.27                                & 25.71                                & 17.81                                \\ \hline
\multirow{4}{*}{\rotatebox[origin=c]{90}{\scalebox{.6}{\sc \textbf{validating}}}} & sentences  &    & 19,233                                & 3,974                                 & 3,054                                 & 2,902                                 & 3,960                                 & 1,770                                 & 8,108                                 \\
                                                                                  & unique tokens & (\%)  & 13.35                              & 16.59                              & 18.56                              & 20.40                              & 32.73                              & 22.65                              & 24.86                              \\
                                                                                  & sentence size & & 17.26                                & 15.76                                & 29.30                                & 19.87                                & 12.07                                & 24.28                                & 17.30                                \\
                                                                                  & unknown tokens & (\%) & 5.46                               & 7.51                               & 5.35                               & 6.41                               & 13.12                              & 5.64                               & 11.69                              \\ \hline
\end{tabular}
\end{center}
\caption{Characteristics of datasets for resource-rich languages}  
\label{tabla-resource-rich-languages-sampling-data-sets}
\end{table}

\begin{table}[]
\begin{center}
\begin{tabular}{lllrrrrrrrr}
\hline
                                                                                  &                      & & \multicolumn{1}{c}{{\bf\scshape ca}} & \multicolumn{1}{c}{{\bf\scshape eu}} & \multicolumn{1}{c}{{\bf\scshape fa}} & \multicolumn{1}{c}{{\bf\scshape gl}} & \multicolumn{1}{c}{{\bf\scshape hi}} & \multicolumn{1}{c}{{\bf\scshape ja}} & \multicolumn{1}{c}{{\bf\scshape sr}} & \multicolumn{1}{c}{{\bf\scshape zh}} \\ \hline
\multirow{3}{*}{\rotatebox[origin=c]{90}{\scalebox{.6}{\sc \textbf{training}}}}   & sentences  &    & 13,123                                & 5,396                                 & 4,798                                 & 2,872                                 & 13,304                                & 7,125                                 & 3,328                                 & 3,997                                 \\
                                                                                  & unique tokens & (\%)  & 7.40                               & 26.34                              & 10.94                              & 15.18                              & 6.01                               & 14.20                              & 22.29                              & 17.78                              \\
                                                                                  & sentence size & & 31.82                                & 13.52                                & 25.23                                & 33.00                                & 21.13                                & 22.48                                & 22.31                                & 24.67                                \\ \hline
\multirow{4}{*}{\rotatebox[origin=c]{90}{\scalebox{.6}{\sc \textbf{validating}}}} & sentences  &    & 1,709                                 & 1,798                                 & 599                                  & \textit{1,260}                        & 1,659                                 & 511                                  & 536                                  & 500                                  \\
                                                                                  & unique tokens & (\%)  & 16.40                              & 36.53                              & 24.50                              & \textit{20.85}                     & 15.16                              & 31.76                              & 37.40                              & 34.00                              \\
                                                                                  & sentence size & & 33.05                                & 13.40                                & 26.43                                & \textit{31.65}                       & 21.23                                & 22.46                                & 22.38                                & 25.33                                \\
                                                                                  & unknown tokens & (\%) & 5.78                               & 19.14                              & 10.46                              & \textit{11.05}                     & 5.51                               & 6.23                               & 18.58                              & 13.03                              \\ \hline
\end{tabular}
\end{center}
\caption{Characteristics of datasets for resource-poor languages}
\label{tabla-resource-poor-languages-sampling-data-sets}
\end{table}

\subsubsection{Software resources}

Analogously to the case of the linguistic resources, we use the most
popular and efficient generators for {\sc dl}-based dependency parsers
from the {\sc n}euro{\sc nlp2} project~\citep{Ma-Hovy-2016}. It
includes implementations for state-of-the-art models for a wide range
of core tasks, including parsing~\citep{Liu-etal-2019, Ma-etal-2018,
  Rotman-Reichart-2019}, {\sc pos} tagging~\citep{Tourille-etal-2018}
and named entity recognition~\citep{Magnini-etal-2020}, all publicly
available online
(\textit{https://github.com/XuezheMax/NeuroNLP2}). Focusing on
parsing, the scope of our case study, we chose to work with two
particular encoders:

\begin{itemize}

\item \textit{Deep biaffine attention} ({\sc b}i{\sc
  af})~\citep{Dozat2017}.

\item \textit{Maximum spanning tree} ({\sc n}euro{\sc
  mst})~\citep{Ma2017}.

\end{itemize}

\noindent This allows, respectively, coverage of projective and
non-projective capabilities~\footnote{The capability to deal with
  \textit{non-projective} tree structures, i.e., that include crossing
  lines, defines a major qualitative classification criterion for
  parsing tools because numerous sentences in many languages require it for
  their syntactic analysis.} when dealing with dependency-based
parsing, precisely the parsing paradigm propelled by the rise of {\sc
  dl} technologies. 

Both encoders share a basic architecture, structured as a recurrent
network using \textit{bidirectional long-short term memory} ({\sc
  b}i{\sc lstm})~\citep{Ma-Hovy-2016} to encode inputs. Both also
operate at word level, although {\sc n}euro{\sc mst} takes the form of
a \textit{convolutional neural network} ({\sc
  cnn})~\citep{Lecun-etal-1998} to exploit information at the
character level. As the main distinctive characteristics of each
encoder, {\sc n}euro{\sc mst}-based parsers rely on a conditional
log-linear model on top of the neural network to classify dependency
labels, while {\sc b}i{\sc af}-based ones do this classification via a
modified version of \textit{bilinear
  attention}~\citep{Kim-etal-2018}. Unlike {\sc n}euro{\sc mst}, {\sc
  b}i{\sc af} uses different vectors for different dependency labels
to represent each word, thus performing slightly better in exchange
for a greater memory requirement. The hyperparameter configuration is
similar for both encoders: word embeddings of 300 --obtained from
\textit{https://fasttext.cc/docs/en/crawl-vectors.html})--
dimensions, 3 recurrent layers of 512 neurons, {\sc elu} activation
units, and a uniform dropout of 0.3.

\subsection{Testing space}

To illustrate how our proposal performs, we need to compare it to other
cross-validation-based online indicators. We also need to ensure an
appropriate setting of parameters so as to introduce the set of local
testing frames that define the testing space.

\subsubsection{State-of-the-art online indicators}

On the basis of the state-of-the-art in both {\sc nn}{\footnotesize s}
and {\sc gp}, we describe the online indicators listed below, as
referred to a given kernel ${\mathcal K}={\mathcal N}^{\pi}[{\mathcal
    D}]$.  In addition to the primary rule
$c_{\textit{mne}[m]}^{\mathcal K}$, which fixes the maximum number $m$
of epochs to be executed, the online indicators and also our proposal,
$c_{\textit{coi}[\aleph,k,\alpha_\aleph]}^{\mathcal K}$, are monitored
against the baseline $c_{\wp[h]}^{\mathcal K}$ provided by an
omniscient oracle in a horizon of $h=10^3$ epochs.

\paragraph{Generalization loss $(c_{\textit{gl}})$}

Training stops when the generalization loss rises above a specific
threshold~\citep{Prechelt97automaticearly} relative to
the optimal error $E_{\textit{op}}[e]$, i.e., the minimal
error observed in epochs up to the current one $e$:
\begin{equation}
  \label{eq-optimal-error-Prechelt}
  E_{\textit{op}}^{\mathcal K}[e] := \min_{\hat{e} \leq e}
  E_{\textit{va}}^{\mathcal K}[\hat{e}]
\end{equation}
\noindent We then define (as a percentage) the loss at epoch $e$ by:
\begin{equation}
  \label{eq-GL-criterion}
  c_{{\textit{gl}}}^{\mathcal K}[e] := 100 \ast
  \left(\frac{E_{\textit{va}}^{\mathcal K}[e]}{E_{\textit{op}}^{\mathcal K}[e]}-1
  \right)
\end{equation}
\noindent and, having fixed a maximum value $\alpha$, the corresponding
online indicator is given by:
\begin{equation}
  \label{eq-GL-alpha-online-indicator}
  c_{{\textit{gl}[\alpha]}}^{\mathcal K}[e] := [c_{{\textit{gl}}}^{\mathcal K}[e] > \alpha]
\end{equation}

\paragraph{Progress $(c_{\textit{p}[k]})$}

Training stops when during a \textit{training strip of length} $k$,
improvements in training error stall below a specific
threshold. Progress is measured as the ratio between the average
training error during the strip and the minimum one. We then define
the progress (in parts per thousand) at epoch $e$ by:
\begin{equation}
  \label{eq-Pk-criterion}
  c_{\textit{p}[k]}^{\mathcal K}[e] := 100 \ast
  \left(\frac{\sum_{\hat{e}=e-k+1}^{e}E_{\textit{tr}}^{\mathcal K}[\hat{e}]}
             {k \ast \min_{\hat{e}=e-k+1}^{e}E_{\textit{tr}}^{\mathcal K}[\hat{e}]}-1
  \right)
\end{equation}
\noindent Having fixed a minimum value $\alpha$, the corresponding
online indicator is given by:
\begin{equation}
  \label{eq-Pk-alpha-online-indicator}
  c_{{\textit{p}[k,\alpha]}}^{\mathcal K}[e] := [c_{\textit{p}[k]}^{\mathcal K}[e] < \alpha]
\end{equation}
\noindent Comparing with $c_{\textit{gl}}$, $c_{{\textit{p}[k]}}$ is
high for unstable phases of training, where the training error goes up
instead of down. This was described by~\cite{Prechelt97automaticearly}
and first evaluated by~\cite{Lodwich:2009:ERP:1704175.1704224} as
\textit{low progress}.

\paragraph{Productivity quotient $(c_{\textit{pq}[k]})$}

Training stops when, during a strip of length $k$, there is little
chance that the generalization loss can be repaired, which may happen
when progress is very rapid. Evaluation is based on a threshold in the
ratio to training progress~\citep{Prechelt97automaticearly}. We then
define the productivity quotient at epoch $e$ as:
\begin{equation}
  \label{eq-PQk-criterion}
  c_{\textit{pq}[k]}^{\mathcal K}[e] := \frac{c_{\textit{gl}}^{\mathcal K}[e]}{c_{\textit{p}[k]}^{\mathcal K}[e]}
\end{equation}
\noindent Having fixed a maximum value $\alpha$, the corresponding
online indicator is given by:
\begin{equation}
  \label{eq-PQk-alpha-online-indicator}
  c_{{\textit{pq}[k,\alpha]}}^{\mathcal K}[e] := [c_{\textit{pq}[k]}^{\mathcal K}[e] > \alpha]
\end{equation}
\noindent Comparing with $c_{\textit{gl}}$, $c_{{\textit{pq}[k]}}$
does not report overfitting until the training error begins to
decrease slowly, which is of interest if we can assume a higher
generalization loss given a greater progress on the training set.

\paragraph{Uninterrupted progress $(c_{\textit{up}[s,k]})$}

Training stops when, during a sequence of $s$ strips of length $k$,
the generalization error
increases~\citep{Prechelt97automaticearly}. We then assume that
overfitting has begun independently of the size of the
increases. Thus, we recursively define uninterrupted progress at
epoch $e$ as:
\begin{equation}
  \label{eq-UPks-online-indicator}
  c_{\textit{up}[s,k]}^{\mathcal K}[e] := [c_{\textit{up}[s-1,k]}^{\mathcal K}[e-k] \; \& \;
  E_{\textit{va}}^{\mathcal K}[e] > E_{\textit{va}}^{\mathcal K}[e-k]], \;
  \forall s \geq 2
\end{equation}
\noindent with
\begin{equation}
  \label{eq-UPk1-online-indicator}
  c_{\textit{up}[1,k]}^{\mathcal K}[e] := [E_{\textit{va}}^{\mathcal K}[e] > E_{\textit{va}}^{\mathcal K}[e-k]]
\end{equation}
\noindent The scope of $c_{{\textit{up}[s,k]}}$ with respect to
the generalization error ($E_{\textit{va}}$) is local while that of
$c_{\textit{gl}}$ is global, because the reference up to the current
epoch $e$ is the minimum ($E_{\textit{op}}$). Thus,
$c_{{\textit{up}[s,k]}}$ can be used directly in the context of
pruning algorithms, where errors are allowed to remain much higher
than previous minima over long training periods.

%

\paragraph{High noise ratio $(c_{\textit{hnr}[k]})$}

Training stops when, during a strip of length $k$, the \textit{high
  noise ratio} ({\sc hnr}) goes above a specific threshold. This
stopping rule is defined~\citep{Lodwich:2009:ERP:1704175.1704224} at
epoch $e$ as:
\begin{equation}
\label{eq-HNRk-criterion}
  c_{\textit{hnr}[k]}^{\mathcal K}[e] := \frac{\sum_{\hat{e}=e-1}^{e-k}E_{\textit{tr}}^{\mathcal K}[\hat{e}] -2 \ast
                                           E_{\textit{tr}}^{\mathcal K}[\hat{e}-1] +
                                           E_{\textit{tr}}^{\mathcal K}[\hat{e}-2]}  
                              {\sum_{\hat{e}=e-1}^{e-k}E_{\textit{tr}}^{\mathcal K}[\hat{e}]}
\end{equation}
\noindent Having fixed a ceiling for $\alpha$, the corresponding
online indicator is given by: 
\begin{equation}
  \label{eq-HNRk-alpha-online-indicator}
  c_{{\textit{hnr}[k,\alpha]}}^{\mathcal K}[e] := [c_{\textit{hnr}[k]}^{\mathcal K}[e] > \alpha]
\end{equation}

\paragraph{Overfitting gain $(c_{\textit{og}})$}

Training stops when the gain in overfitting, with respect to the
minimal gain observed in epochs up to the current epoch, rises above a
specific threshold. In accordance with the measurement of overfitting
in {\sc gp} from~\cite{Vanneschi:2010:MBO:1830483.1830643}, this gain
is interpreted as the distance between training and validation
errors. Formally, we define the overfitting gain at epoch $e$ as:
\begin{equation}
\label{eq-OG-Vanneschi-like}
  \textit{Og}^{\mathcal K}[e] := \abs{E_{\textit{tr}}^{\mathcal K}[e] - E_{\textit{va}}^{\mathcal K}[e]}
\end{equation}
\noindent from which its optimal value in epochs up to $e$ is given by:
\begin{equation}
  \label{eq-optimal-OG-Vanneschi-like}
    \textit{Og}_{\textit{op}}^{\mathcal K}[e] := \min_{\hat{e} \leq e} \textit{Og}^{\mathcal K}[\hat{e}] 
\end{equation}
\noindent We can define the stopping criterion for this epoch as:
\begin{equation}
\label{eq-OG-criterion}
c_{\textit{og}}^{\mathcal K}[e] := \abs{\textit{Og}^{\mathcal K}[e]}- \abs{\textit{Og}_{\textit{op}}^{\mathcal K}[e]} 
\end{equation}
\noindent Having fixed a ceiling for $\alpha$, the corresponding online
indicator is given by:
\begin{equation}
  \label{eq-OG-Vanneschi-like-indicator}
  c_{{\textit{og}[\alpha]}}^{\mathcal K}[e] := [c_{\textit{og}}^{\mathcal K}[e] > \alpha]
\end{equation}

\subsubsection{Parameter tuning}

The performance of online indicators relies heavily on a study of the
{\sc ml} process, especially when they include parameters
that need to be tuned in order to obtain the best results and thus
ensure reliable evaluation, which in turn requires properly setting
the quality metrics. In our case, the latter refers to the parameter
fitting stage in \nemesid{}, through the functions used
to estimate the cost and error of the runs.

\paragraph{Setting the \nemesid{} metric}

Given a run ${\mathcal R} = {\mathcal N}^{\pi}_{c_\iota}[{\mathcal
    D}]$ for a sampling database ${\mathcal D} = {\mathcal
  D}_{\textit{te}} \cup {\mathcal D}_{\textit{tr}} \cup {\mathcal
  D}_{\textit{va}}$ and an online indicator ${c_\iota}$, we first
define the loss function $E$ to be used to measure the error in one of
the sets ${\mathcal D}_j \subseteq {\mathcal D}, \; j \in
\{\textit{te}, {\textit{tr}}, {\textit{va}}\}$. To facilitate
interpretation of the results, we opt for conceptual transparency,
with the error measured as the simple counterpart of the accuracy
($\accuracy{{\mathcal R}}{j}$) achieved by the run for the set, as
follows:
\begin{equation}
\label{eq-loss-function-testing-frame}
E_j^{\mathcal R} := 100 - \accuracy{{\mathcal R}}{j}
\end{equation}
\noindent For its part, accuracy is interpreted in the more usual sense of the
state-of-the-art~\citep{Sampson2004corpus}, i.e., as the number of
correctly identified parse trees divided by the total number of parse
trees (expressed as a percentage), calculated following a standard
procedure. So, all parses in the database ${\mathcal D}_j$ are counted and
it is assumed that only a single parse \textit{per} sentence is
provided. Considering that we are working with dependency parsers, we
can formalize this procedure as:
\begin{equation}
\label{eq-accuracy-dependency-parsing}
\accuracy{{\mathcal R}}{j} := 100 * \frac{\abs{\pheads{{\mathcal R}}{j} \cap \heads{j}}}{\abs{\heads{j}}}
\end{equation}
\noindent where $\heads{j}$ is the set of heads, i.e., the words starting a syntactic dependency, in database ${\mathcal D}_j$, and $\pheads{{\mathcal R}}{j}$ is the set of heads predicted by the model ${\mathcal R}$ for the same
database.

Regarding the cost function, depending on the neural architecture and
sampling database ${\mathcal D}$ associated with run ${\mathcal R}$,
it may be convenient to increase or decrease the weighting factors
$\omega_{\textit{mi}}$ and $\omega_{\textit{ea}}$, referring to the
costs of executing and validating errors in the model, respectively.
However, when linguistic and software resources of a
very different nature are involved, as in our case, we cannot
standardize them through a collection of local testing
frames. We therefore work with $\omega_{\textit{mi}} =
\omega_{\textit{ea}} = 0.5$, a couple of intermediate values in the
range of possible ones. In practice, this avoids extreme weights
and, therefore, any misconceptions or inaccurate conclusions regarding
the tests.

\paragraph{Setting the online indicators}

The aim is to identify the best performance setting for each online
indicator in each particular run. Depending on the case, this means
calculating three types of values: upper/lower threshold $\alpha$,
training strip length $k$ and strip sequence length $s$. To do this,
we tune those values on the horizon $h = 10^3$ of the kernel
considered ${\mathcal K}={\mathcal N}^{\pi}[{\mathcal D}]$, taking the
score of the \nemesid{} metric for the optimal online indicator
$c_{\wp[h]}^{\mathcal K}$ as a benchmark and increasing values in
intervals proposed by the state-of-the-art. Thus:

\begin{itemize}

\item For the rule of fixing a \textit{maximum number of epochs}, we
  tune $m$ in the condition $c_{\textit{mne}[m]}^{\mathcal K}$
  in increments of $10$ over the interval $[10, 100]$ proposed
  by both~\cite{Vanneschi:2010:MBO:1830483.1830643} and~\cite{Rosasco15}.

  Since the horizon $h$ considered for the oracle far exceeds this
  interval, this tuning does not guarantee optimal performance, but
  only allows us to approximate the best possible one in the range of
  epochs usually considered. Note that if we were to complete the
  estimation at $[0,h]$, we would virtually reproduce the behavior of
  the oracle, thus eliminating the practical interest of a comparison.

\item  For the \textit{generalization loss}, we tune $\alpha$
  for the online indicator $c_{\textit{gl}[\alpha]}^{\mathcal K}$ in
  increments of $0.5$ over the interval $[1,5]$ proposed
  by~\cite{Prechelt97automaticearly}.

\item For \textit{progress}, we take $k=5$ and tune
  $\alpha$ for the online indicator
  $c_{\textit{p}[k,\alpha]}^{\mathcal K}$ in increments of $0.5$ over
  the interval $[1,5]$ proposed
  by~\cite{Lodwich:2009:ERP:1704175.1704224}.

\item For the \textit{productivity quotient}, we take
  $k=5$ and tune $\alpha$ for the online indicator
  $c_{\textit{pq}[k,\alpha]}^{\mathcal K}$ in increments of $0.5$
  over the interval $[1,5]$ proposed by~\cite{Prechelt97automaticearly}.

\item For \textit{uninterrupted progress}, we take $k=s=5$ 
  for the online indicator $c_{\textit{up}[k,s]}^{\mathcal R}$, as
  proposed by~\cite{Prechelt97automaticearly}. 

\item For the \textit{high noise ratio}, we take $k=5$ and tune
  $\alpha$ for the online indicator
  $c_{\textit{hnr}[k,\alpha]}^{\mathcal K}$ in increments of $0.5$
  over the interval $[5,25]$ proposed
  by~\cite{Lodwich:2009:ERP:1704175.1704224}.

\item For the \textit{overfitting gain}, we tune $\alpha$ for the
  online indicator $c_{\textit{og}[\alpha]}^{\mathcal K}$ in
  increments of $0.5$ over the interval $[0.5,5]$ proposed
  by~\cite{Vanneschi:2010:MBO:1830483.1830643}.
  
\end{itemize}

\noindent For our own proposal
$c_{\textit{coi}[\aleph,k,\alpha_\aleph]}^{\mathcal K}$ and regarding
the training strip length $k$, we make the same choice ($k=5$) as for
the other indicators using that parameter. Turning to the value for
the lower threshold $\alpha_\aleph$, it is tuned in increments of
$0.1$ over the interval $[0.5,1]$, which also relates to a selection
procedure analogous to the one applied for the rest of the stopping
conditions depending of this kind of setting ($\alpha$). When it comes
to the correlation coefficient $\aleph$, it is a parameter specific to
the new indicator. Consequently, our decision is only conditioned by
the nature of the values to be related, finally falling to the
\textit{Pearson's} $\rho$~\citep{Pearson1895}, the popular method to
measure the linear correlation between two sets of normally
distributed data. To justify it, one need only recall the {\em Central
  Limit Theorem}~\citep{Polya-1920}, which proves that a stochastic
variable depending on a large number of small effects, tends to
approximate a normal distribution. This is typically the case for
observation error calculations and, in particular, for the online
indicators whose correlation underpins our proposal, thus giving
formal support to our choice. In any case, and to give a broader view
of the impact of this choice, we have performed the same
tests using the {\em Spearman's rank order}~\citep{Spearman1904}, a
general correlation coefficient that evaluates monotonic
relationships, whether linear or non-linear. The results obtained from
these alternative correlation values have not differed from those
associated with Pearson's $\rho$, which is why we do not make an
express distinction between the two coefficients.

On the whole, the settings applied do not, therefore, represent in any
way an advantage over our competitors in the design of the testing
space, which is a major factor in the credibility of any positive
evaluation of the new stopping condition described.

\subsubsection{Local testing frames}

Once the cross-validation has been set as canary function $c$ around
which to define our testing frame, the correlation coefficient
$\aleph$ selected, and the settings for the collection $\Upsilon_c$ of
most representative associated indicators have been optimized, we can
formally introduce our experimental space. This involves
characterizing our collection of local testing frames, whose aim is to
categorize the runs to be studied, using the concept of a kernel,
i.e. a learning task involving a sampling database and a
learner. Thus, the family $\Upsilon_c$ of online indicators to be
considered is:
\begin{equation}
\label{eq-collection-online-indicators-testing-frame}
  \Upsilon_c := \{ c_{\textit{coi}[\aleph,k,\alpha_\aleph]},
  c_{\textit{gl}[\alpha]},
  c_{{\textit{hnr}[k,\alpha]}},
  c_{\textit{mne}[m]},
  c_{{\textit{og}[\alpha]}},
  c_{\textit{p}[k, \alpha]},
  c_{{\textit{pq}[k,\alpha]}},
  c_{{\textit{up}[s,k]}}\}
\end{equation}

\noindent All of these, except for $c_{\textit{mne}[m]}$, will operate
in association with a primary rule $c_{\textit{mne}[h]}$, which
guarantees that all runs stop at some point within the limits of the
horizon $h$ set for the oracle. With this aim, we will say that a
run is \textit{out-of-range} when the stop is produced by
activation of that primary rule. \\

\noindent The {\sc dl} encoders {\sc b}i{\sc af} and {\sc n}euro{\sc
  mst} and the collection of {\sc ud} treebanks of resource-rich
(resp. resource-poor) languages are:
\begin{equation}
\label{eq-UD-treebanks}
  R := \{ \textsc{zh, en, fa, fr, de, hi, ja,
  pl, pt, ru, es}\} \mbox{ (resp. }
P := \{\textsc{ca, eu, gl, sr} \}\mbox{)}
\end{equation}
\noindent The local testing frame families to be considered are
therefore the following ones:
\begin{equation}
\label{eq-families-local-testing-frames-1}
  \{ \mathcal{L}_{R,d}^{\Upsilon_c}\}_{d \in {\mathcal D}} \mbox{ (resp. }
 \{ \mathcal{L}_{P,d}^{\Upsilon_c}\}_{d \in {\mathcal D}}\mbox{)}, \mbox{ with } {\mathcal D} := \{\mbox{{\sc b}i{\sc af}}, \mbox{{\sc n}euro{\sc mst}}\}
\end{equation}

\noindent one \textit{per} combination of corpus in $R$ (resp. $P$)
and encoder $d \in {\mathcal D}$ set, applying in all its runs the
parameter range estimated above. These four sets, $\{
\mathcal{L}_{R,d}^{\Upsilon_c}\}_{d \in {\mathcal D}}$ and $\{
\mathcal{L}_{P,d}^{\Upsilon_c}\}_{d \in {\mathcal D}}$, together with
the \nemesid{} quality metric make up our testing frame.

\subsection{Results analysis}

The rows in
Tables~\ref{table-monitoring-local-testing-frames-biaf-resource-rich-languages}
and~\ref{table-monitoring-local-testing-frames-neuromst-resource-rich-languages}
(resp. Tables~\ref{table-monitoring-local-testing-frames-biaf-resource-poor-languages}
and~\ref{table-monitoring-local-testing-frames-neuromst-resource-poor-languages})
show monitoring details for each local testing frame in $\{
\mathcal{L}_{{\mathcal R},d}^{\Upsilon_c}\}_{d \in {\mathcal D}}$
(resp.  $\{ \mathcal{L}_{{\mathcal P},d}^{\Upsilon_c}\}_{d \in
  {\mathcal D}}$). The entries describe the runs associated with
online indicators $c_i$ in the considered collection $\Upsilon_c$,
referring to the parameter setting for $c_i$. This latter includes the
epoch $s_{c_i}$ at which possible overfitting occurs when the
indicator is applied in the corresponding {\em corpus}, and the value
computed by the \nemesid{} function
$\Phi^{\omega_{mi}}_{\omega_{\textit{ea}}}$. Within a local testing
frame, \textit{italics} indicates the baseline results and
\textbf{bold} indicates the best performance with respect to the
baseline, always without taking out-of-range runs into account. In
order to simplify their interpretation, these tables are analyzed with
the aid of classic bar charts in
Figs.~\ref{fig-percentages-ranking-positions-on-biaf-neuromst-resource-rich-languages}
and~\ref{fig-percentages-ranking-positions-on-biaf-neuromst-resource-poor-languages},
and also by means of box plots when it comes to comparing
distributions across different datasets in
Figs.~\ref{fig-nemesid-box-plots-on-biaf-neuromst-resource-rich-languages}
and~\ref{fig-nemesid-box-plots-on-biaf-neuromst-resource-poor-languages}. To
improve the understanding of these box plots, we show not only
outliers ($\bullet$) and extreme values (\ding{84}) but also the rest
of the observations ($\blacktriangle$), including now those
corresponding to out-of-range runs, for which we use the same symbols
in a lighter color (\textcolor{lightgray}{$\bullet$},
\textcolor{lightgray}{\ding{84}},
\textcolor{lightgray}{$\blacktriangle$}). Further, we associate the
mean ($\mu$) and variance ($\sigma^2$) achieved by each online
indicator for the set of runs considered. All numerical data are
expressed to two decimal places in order to improve readability, while
all the calculations have been done to ten decimal places of
precision.

%
%

As pointed out, \nemesid{} is an estimate, in {\sc meu} terms, of the
signed and normalized cost deviation of a run with respect to the baseline in
its local testing frame. Since it is a metric whose codomain is
$[-1, 1]$ and whose value (resp. modulus) is
(resp. inversely) proportional to computational cost (resp. utility),
we hope to reach rates with the lowest absolute values, and preferably
negative ones in non-null cases.

\subsubsection{The quantitative study}

We are now interested in analyzing the behavior of our proposal
$c_{\textit{coi}[\aleph,5,\alpha_\aleph]}$ from both a relative and an
absolute point of view, i.e., with respect to other online indicators
in the state-of-the-art, but also to the oracle $c_{\wp[h]}, \; h=10^3$.

\begin{table}[htbp]
\centering
\resizebox{\textwidth}{!}{%
\tabcolsep=0.1cm
\begin{tabular}{rrrrrrrrrrrrrrrrrrrrrrrrrrrrrrrrrrrrrrrrrrrrrrr}
\hline
                 &  & \multicolumn{3}{c}{$c_{\wp[10^3]}$}                                              &                      & \multicolumn{5}{c}{\color{violet}$c_{\textit{coi}[\aleph,5,\alpha_{\aleph}]}$} &                      & \multicolumn{5}{c}{$c_{{\textit{gl}[\alpha]}}$}                                              &                      & \multicolumn{3}{c}{$c_{\textit{hnr}[5,5.0]}$}                                              &                      & \multicolumn{5}{c}{$c_{\textit{mne}[m]}$}                                                                        &                      & \multicolumn{5}{c}{$c_{\textit{og}[\alpha]}$}                                                                             &                      & \multicolumn{3}{c}{$c_{{\textit{p}[5,1.0]}}$}                                              &                      & \multicolumn{5}{c}{$c_{\textit{pq}[5,\alpha]}$}                                                                             &                      & \multicolumn{3}{c}{$c_{\textit{up}[5,5]}$}      \\                                         \cline{1-1} \cline{3-5} \cline{7-11} \cline{13-17} \cline{19-21} \cline{23-27} \cline{29-33} \cline{35-37} \cline{39-43} \cline{45-47}
                 &  & \multicolumn{1}{c}{$s_{c_{\wp[10^3]}}$} & & \multicolumn{1}{c}{$\Phi_{0.5}^{0.5}$} & \multicolumn{1}{c}{} & \multicolumn{1}{c}{\color{violet}$\alpha_{\aleph}$} & & \multicolumn{1}{c}{\color{violet}$s_{c_{\textit{coi}[\aleph,5,\alpha_{\aleph}]}}$} & & \multicolumn{1}{c}{\color{violet}$\Phi_{0.5}^{0.5}$} & \multicolumn{1}{c}{} & \multicolumn{1}{c}{$\alpha$} & \multicolumn{1}{c}{} & \multicolumn{1}{c}{$s_{c_{{\textit{gl}[\alpha]}}}$} & & \multicolumn{1}{c}{$\Phi_{0.5}^{0.5}$} & \multicolumn{1}{c}{} & \multicolumn{1}{c}{$s_{c_{\textit{hnr}[5,5.0]}}$} & & \multicolumn{1}{c}{$\Phi_{0.5}^{0.5}$} & \multicolumn{1}{c}{} & \multicolumn{1}{c}{$m$} & & \multicolumn{1}{c}{$s_{c_{\textit{mne}[m]}}$} & & \multicolumn{1}{c}{$\Phi_{0.5}^{0.5}$} & \multicolumn{1}{c}{} & \multicolumn{1}{c}{$\alpha$} & & \multicolumn{1}{c}{$s_{c_{\textit{og}[\alpha]}}$} & & \multicolumn{1}{c}{$\Phi_{0.5}^{0.5}$} & \multicolumn{1}{c}{} & \multicolumn{1}{c}{$s_{c_{{\textit{p}[5,1.0]}}}$} & & \multicolumn{1}{c}{$\Phi_{0.5}^{0.5}$} & \multicolumn{1}{c}{} & \multicolumn{1}{c}{$\alpha$} & & \multicolumn{1}{c}{$s_{c_{\textit{pq}[5,\alpha]}}$} & & \multicolumn{1}{c}{$\Phi_{0.5}^{0.5}$} & \multicolumn{1}{c}{} & \multicolumn{1}{c}{$s_{c_{\textit{up}[5,5]}}$} & & \multicolumn{1}{c}{$\Phi_{0.5}^{0.5}$} \\ \cline{1-1} \cline{3-3} \cline{5-5} \cline{7-7} \cline{9-9} \cline{11-11} \cline{13-13} \cline{15-15} \cline{17-17} \cline{19-19} \cline{21-21} \cline{23-23} \cline{25-25} \cline{27-27} \cline{29-29}  \cline{31-31} \cline{33-33} \cline{35-35} \cline{37-37} \cline{39-39} \cline{41-41} \cline{43-43} \cline{45-45} \cline{47-47}

{\bf\scshape de} & & \textit{206} & & \textit{0.00} & & \color{violet} 0.5 & & \color{violet} 88 & & \color{violet} -0.15 & & 1.0 & & 5 & & -0.25 & & $10^3$ & & 1.00 & & 90 & & 90 & & -0.15 & & 0.5 & & 8 & & -0.25 & & \textbf{284} & & \textbf{0.10} & & 1.0 & & \textbf{284} & & \textbf{0.10} & & $10^3$ & & 1.00 \\
{\bf\scshape en} & & \textit{41} & & \textit{0.00} & & \color{violet} 0.5 & & \color{violet} \textbf{41} & & \color{violet} \textbf{0.00} & & 1.0 & & 4 & & -0.03 & & 226 & & 0.19 & & 40 & & 40 & & -0.00 & & 4.5 & & 32 & & -0.01 & & 67 & & 0.03 & & 1.0 & & 67 & & 0.03 & & $10^3$ & & 1.00 \\
{\bf\scshape es} & & \textit{42} & & \textit{0.00} & & \color{violet} 0.5 & & \color{violet} \textbf{42} & & \color{violet} \textbf{0.00} & & 1.0 & & 6 & & -0.04 & & 135 & & 0.10 & & 40 & & 40 & & -0.00 & & 3.0 & & 39 & & -0.00 & & 96 & & 0.06 & & 1.0 & & 96 & & 0.06 & & 109 & & 0.07 \\
{\bf\scshape fr} & & \textit{4} & & \textit{0.00} & & \color{violet} 0.5 & & \color{violet} 5 & & \color{violet} 0.00 & & 1.0 & & \textbf{1} & & \textbf{0.00} & & 168 & & 0.17 & & 10 & & 10 & & 0.01 & & 0.5 & & \textbf{1} & & \textbf{0.00} & & 45 & & 0.04 & & 1.0 & & 45 & & 0.04 & & 208 & & 0.21 \\
{\bf\scshape pl} & & \textit{22} & & \textit{0.00} & & \color{violet} 0.5 & & \color{violet} \textbf{21} & & \color{violet} \textbf{-0.00} & & 1.0 & & 2 & & -0.02 & & 111 & & 0.09 & & 20 & & 20 & & -0.00 & & 1.0 & & 11 & & -0.01 & & 37 & & 0.02 & & 1.0 & & 37 & & 0.02 & & $10^3$ & & 1.00 \\
{\bf\scshape pt} & & \textit{27} & & \textit{0.00} & & \color{violet} 0.7 & & \color{violet} \textbf{29} & & \color{violet} \textbf{0.00} & & 1.0 & & 7 & & -0.02 & & 92 & & 0.07 & & 30 & & 30 & & 0.00 & & 0.5 & & 2 & & -0.02 & & $10^3$ & & 1.00 & & 1.0 & & $10^3$ & & 1.00 & & $10^3$ & & 1.00 \\
{\bf\scshape ru} & & \textit{14} & & \textit{0.00} & & \color{violet} 0.5 & & \color{violet} 16 & & \color{violet} 0.00 & & 1.0 & & 6 & & -0.01 & & 60 & & 0.05 & & 10 & & 10 & & -0.00 & & 3.0 & & \textbf{13} & & \textbf{-0.00} & & 40 & & 0.03 & & 1.0 & & 40 & & 0.03 & & $10^3$ & & 1.00 \\
\hline
\end{tabular}
}
\caption{ {\sc b}i{\sc af} encoder monitoring of local testing frames for resource-rich languages}
\label{table-monitoring-local-testing-frames-biaf-resource-rich-languages}
\end{table}

\begin{table}[htbp]
\centering
\resizebox{\textwidth}{!}{%
\tabcolsep=0.1cm
\begin{tabular}{rrrrrrrrrrrrrrrrrrrrrrrrrrrrrrrrrrrrrrrrrrrrrrr}
\hline
                 &  & \multicolumn{3}{c}{$c_{\wp[10^3]}$}                                              &                      & \multicolumn{5}{c}{\color{violet}$c_{\textit{coi}[\aleph,5,\alpha_{\aleph}]}$} &                      & \multicolumn{5}{c}{$c_{{\textit{gl}[\alpha]}}$}                                              &                      & \multicolumn{3}{c}{$c_{\textit{hnr}[5,5.0]}$}                                              &                      & \multicolumn{5}{c}{$c_{\textit{mne}[m]}$}                                                                        &                      & \multicolumn{5}{c}{$c_{\textit{og}[\alpha]}$}                                                                             &                      & \multicolumn{3}{c}{$c_{{\textit{p}[5,1.0]}}$}                                              &                      & \multicolumn{5}{c}{$c_{\textit{pq}[5,\alpha]}$}                                                                             &                      & \multicolumn{3}{c}{$c_{\textit{up}[5,5]}$}      \\                                         \cline{1-1} \cline{3-5} \cline{7-11} \cline{13-17} \cline{19-21} \cline{23-27} \cline{29-33} \cline{35-37} \cline{39-43} \cline{45-47}
                 &  & \multicolumn{1}{c}{$s_{c_{\wp[10^3]}}$} & & \multicolumn{1}{c}{$\Phi_{0.5}^{0.5}$} & \multicolumn{1}{c}{} & \multicolumn{1}{c}{\color{violet}$\alpha_{\aleph}$} & & \multicolumn{1}{c}{\color{violet}$s_{c_{\textit{coi}[\aleph,5,\alpha_{\aleph}]}}$} & & \multicolumn{1}{c}{\color{violet}$\Phi_{0.5}^{0.5}$} & \multicolumn{1}{c}{} & \multicolumn{1}{c}{$\alpha$} & \multicolumn{1}{c}{} & \multicolumn{1}{c}{$s_{c_{{\textit{gl}[\alpha]}}}$} & & \multicolumn{1}{c}{$\Phi_{0.5}^{0.5}$} & \multicolumn{1}{c}{} & \multicolumn{1}{c}{$s_{c_{\textit{hnr}[5,5.0]}}$} & & \multicolumn{1}{c}{$\Phi_{0.5}^{0.5}$} & \multicolumn{1}{c}{} & \multicolumn{1}{c}{$m$} & & \multicolumn{1}{c}{$s_{c_{\textit{mne}[m]}}$} & & \multicolumn{1}{c}{$\Phi_{0.5}^{0.5}$} & \multicolumn{1}{c}{} & \multicolumn{1}{c}{$\alpha$} & & \multicolumn{1}{c}{$s_{c_{\textit{og}[\alpha]}}$} & & \multicolumn{1}{c}{$\Phi_{0.5}^{0.5}$} & \multicolumn{1}{c}{} & \multicolumn{1}{c}{$s_{c_{{\textit{p}[5,1.0]}}}$} & & \multicolumn{1}{c}{$\Phi_{0.5}^{0.5}$} & \multicolumn{1}{c}{} & \multicolumn{1}{c}{$\alpha$} & & \multicolumn{1}{c}{$s_{c_{\textit{pq}[5,\alpha]}}$} & & \multicolumn{1}{c}{$\Phi_{0.5}^{0.5}$} & \multicolumn{1}{c}{} & \multicolumn{1}{c}{$s_{c_{\textit{up}[5,5]}}$} & & \multicolumn{1}{c}{$\Phi_{0.5}^{0.5}$} \\ \cline{1-1} \cline{3-3} \cline{5-5} \cline{7-7} \cline{9-9} \cline{11-11} \cline{13-13} \cline{15-15} \cline{17-17} \cline{19-19} \cline{21-21} \cline{23-23} \cline{25-25} \cline{27-27} \cline{29-29}  \cline{31-31} \cline{33-33} \cline{35-35} \cline{37-37} \cline{39-39} \cline{41-41} \cline{43-43} \cline{45-45} \cline{47-47}          
{\bf\scshape ca} & & \textit{64} & & \textit{0.00} & & \color{violet} 0.7 & & \color{violet} 55 & & \color{violet} -0.01 & & 2.5 & & 3 & & -0.06 & & 101 & & 0.04 & & 60 & & \textbf{60} & & \textbf{-0.00} & & 2.0 & & 69 & & 0.01 & & $10^3$ & & 1.00 & & 1.0 & & $10^3$ & & 1.00 & & $10^3$ & & 1.00 \\
{\bf\scshape eu} & & \textit{572} & & \textit{0.00} & & \color{violet} 1.0 & & \color{violet} $10^3$ & & \color{violet} 0.76 & & 1.0 & & 2 & & -1.00 & & 86 & & -0.86 & & 90 & & 90 & & -0.85 & & 3.0 & & 6 & & -0.99 & & $10^3$ & & 0.76 & & 1.0 & & $10^3$ & & 0.76 & & \textbf{194} & & \textbf{-0.67} \\
{\bf\scshape fa} & & \textit{864} & & \textit{0.00} & & \color{violet} 1.0 & & \color{violet} $10^3$ & & \color{violet} 0.16 & & 1.0 & & 6 & & -1.00 & & 75 & & -0.92 & & 90 & & \textbf{90} & & \textbf{-0.90} & & 5.0 & & 27 & & -0.98 & & $10^3$ & & 0.16 & & 1.5 & & $10^3$ & & 0.16 & & $10^3$ & & 0.16 \\
{\bf\scshape gl} & & \textit{20} & & \textit{0.00} & & \color{violet} 0.5 & & \color{violet} 19 & & \color{violet} 0.00 & & 1.0 & & 5 & & -0.01 & & 95 & & 0.08 & & 20 & & \textbf{20} & & \textbf{0.00} & & 4.5 & & 19 & & 0.00 & & 81 & & 0.06 & & 1.0 & & 81 & & 0.06 & & $10^3$ & & 1.00 \\
{\bf\scshape hi} & & \textit{18} & & \textit{0.00} & & \color{violet} 0.5 & & \color{violet} 13 & & \color{violet} -0.00 & & 1.0 & & 6 & & -0.01 & & 59 & & 0.04 & & 20 & & \textbf{20} & & \textbf{0.00} & & 1.0 & & 22 & & 0.00 & & $10^3$ & & 1.00 & & 1.0 & & $10^3$ & & 1.00 & & $10^3$ & & 1.00 \\
{\bf\scshape ja} & & \textit{4} & & \textit{0.00} & & \color{violet} 1.0 & & \color{violet} $10^3$ & & \color{violet} 1.00 & & 1.0 & & \textbf{2} & & \textbf{-0.00} & & 15 & & 0.01 & & 10 & & 10 & & 0.01 & & 0.5 & & $10^3$ & & 1.00 & & 8 & & 0.00 & & 1.0 & & 5 & & 0.00 & & $10^3$ & & 1.00 \\
{\bf\scshape sr} & & \textit{284} & & \textit{0.00} & & \color{violet} 0.5 & & \color{violet} \textbf{92} & & \color{violet} \textbf{-0.27} & & 1.0 & & 2 & & -0.38 & & 59 & & -0.31 & & 90 & & 90 & & -0.27 & & 4.5 & & 23 & & -0.36 & & $10^3$ & & 1.00 & & 1.0 & & $10^3$ & & 1.00 & & $10^3$ & & 1.00 \\
{\bf\scshape zh} & & \textit{5} & & \textit{0.00} & & \color{violet} 0.5 & & \color{violet} \textbf{5} & & \color{violet} \textbf{0.00} & & 1.0 & & 3 & & 0.01 & & 64 & & 0.07 & & 10 & & 10 & & 0.01 & & 4.0 & & 4 & & 0.01 & & $10^3$ & & 1.00 & & 1.0 & & 42 & & 0.05 & & 524 & & 0.53 \\
\hline
\end{tabular}%
}
\caption{ {\sc b}i{\sc af} encoder monitoring of local testing frames for resource-poor languages}
\label{table-monitoring-local-testing-frames-biaf-resource-poor-languages}
\end{table}


\begin{table}[htbp]
\centering
\resizebox{\textwidth}{!}{%
\tabcolsep=0.1cm
\begin{tabular}{rrrrrrrrrrrrrrrrrrrrrrrrrrrrrrrrrrrrrrrrrrrrrrr}
\hline
                 &  & \multicolumn{3}{c}{$c_{\wp[10^3]}$}                                              &                      & \multicolumn{5}{c}{\color{violet}$c_{\textit{coi}[\aleph,5,\alpha_{\aleph}]}$} &                      & \multicolumn{5}{c}{$c_{{\textit{gl}[\alpha]}}$}                                              &                      & \multicolumn{3}{c}{$c_{\textit{hnr}[5,5.0]}$}                                              &                      & \multicolumn{5}{c}{$c_{\textit{mne}[m]}$}                                                                        &                      & \multicolumn{5}{c}{$c_{\textit{og}[\alpha]}$}                                                                             &                      & \multicolumn{3}{c}{$c_{{\textit{p}[5,1.0]}}$}                                              &                      & \multicolumn{5}{c}{$c_{\textit{pq}[5,\alpha]}$}                                                                             &                      & \multicolumn{3}{c}{$c_{\textit{up}[5,5]}$}      \\                                         \cline{1-1} \cline{3-5} \cline{7-11} \cline{13-17} \cline{19-21} \cline{23-27} \cline{29-33} \cline{35-37} \cline{39-43} \cline{45-47}
                 &  & \multicolumn{1}{c}{$s_{c_{\wp[10^3]}}$} & & \multicolumn{1}{c}{$\Phi_{0.5}^{0.5}$} & \multicolumn{1}{c}{} & \multicolumn{1}{c}{\color{violet}$\alpha_{\aleph}$} & & \multicolumn{1}{c}{\color{violet}$s_{c_{\textit{coi}[\aleph,5,\alpha_{\aleph}]}}$} & & \multicolumn{1}{c}{\color{violet}$\Phi_{0.5}^{0.5}$} & \multicolumn{1}{c}{} & \multicolumn{1}{c}{$\alpha$} & \multicolumn{1}{c}{} & \multicolumn{1}{c}{$s_{c_{{\textit{gl}[\alpha]}}}$} & & \multicolumn{1}{c}{$\Phi_{0.5}^{0.5}$} & \multicolumn{1}{c}{} & \multicolumn{1}{c}{$s_{c_{\textit{hnr}[5,5.0]}}$} & & \multicolumn{1}{c}{$\Phi_{0.5}^{0.5}$} & \multicolumn{1}{c}{} & \multicolumn{1}{c}{$m$} & & \multicolumn{1}{c}{$s_{c_{\textit{mne}[m]}}$} & & \multicolumn{1}{c}{$\Phi_{0.5}^{0.5}$} & \multicolumn{1}{c}{} & \multicolumn{1}{c}{$\alpha$} & & \multicolumn{1}{c}{$s_{c_{\textit{og}[\alpha]}}$} & & \multicolumn{1}{c}{$\Phi_{0.5}^{0.5}$} & \multicolumn{1}{c}{} & \multicolumn{1}{c}{$s_{c_{{\textit{p}[5,1.0]}}}$} & & \multicolumn{1}{c}{$\Phi_{0.5}^{0.5}$} & \multicolumn{1}{c}{} & \multicolumn{1}{c}{$\alpha$} & & \multicolumn{1}{c}{$s_{c_{\textit{pq}[5,\alpha]}}$} & & \multicolumn{1}{c}{$\Phi_{0.5}^{0.5}$} & \multicolumn{1}{c}{} & \multicolumn{1}{c}{$s_{c_{\textit{up}[5,5]}}$} & & \multicolumn{1}{c}{$\Phi_{0.5}^{0.5}$} \\ \cline{1-1} \cline{3-3} \cline{5-5} \cline{7-7} \cline{9-9} \cline{11-11} \cline{13-13} \cline{15-15} \cline{17-17} \cline{19-19} \cline{21-21} \cline{23-23} \cline{25-25} \cline{27-27} \cline{29-29}  \cline{31-31} \cline{33-33} \cline{35-35} \cline{37-37} \cline{39-39} \cline{41-41} \cline{43-43} \cline{45-45} \cline{47-47}   
{\bf\scshape de} & & \textit{66} & & \textit{0.00} & & \color{violet} 0.5 & & \color{violet} 61 & & \color{violet} -0.01 & & 2.0 & & 4 & & -0.06 & & $10^3$ & & 1.00 & & 70 & & \textbf{70} & & \textbf{0.00} & & 0.5 & & 4 & & -0.06 & & 215 & & 0.16 & & 1.0 & & 215 & & 0.16 & & $10^3$ & & 1.00 \\
{\bf\scshape en} & & \textit{334} & & \textit{0.00} & & \color{violet} 5.0 & & \color{violet} \textbf{345} & & \color{violet} \textbf{0.02} & & 1.0 & & 5 & & -0.49 & & 202 & & -0.20 & & 90 & & 90 & & -0.37 & & 5.0 & & 19 & & -0.47 & & 114 & & -0.33 & & 1.0 & & 114 & & -0.33 & & 72 & & -0.39 \\
{\bf\scshape es} & & \textit{71} & & \textit{0.00} & & \color{violet} 0.5 & & \color{violet} \textbf{70} & & \color{violet} \textbf{-0.00} & & 3.5 & & 19 & & -0.05 & & 101 & & 0.03 & & 70 & & \textbf{70} & & \textbf{-0.00} & & 3.0 & & 58 & & -0.01 & & $10^3$ & & 1.00 & & 1.0 & & $10^3$ & & 1.00 & & 873 & & 0.86 \\
{\bf\scshape fr} & & \textit{230} & & \textit{0.00} & & \color{violet} 1.0 & & \color{violet} \textbf{208} & & \color{violet} \textbf{-0.03} & & 3.5 & & 12 & & -0.28 & & 113 & & -0.15 & & 90 & & 90 & & -0.18 & & 2.0 & & 90 & & -0.18 & & 191 & & -0.05 & & 1.0 & & 191 & & -0.05 & & 255 & & 0.03 \\
{\bf\scshape pl} & & \textit{103} & & \textit{0.00} & & \color{violet} 1.0 & & \color{violet} 85 & & \color{violet} -0.02 & & 3.5 & & 9 & & -0.10 & & 183 & & 0.09 & & 90 & & \textbf{90} & & \textbf{-0.01} & & 2.0 & & 84 & & -0.02 & & 82 & & -0.02 & & 1.0 & & 82 & & -0.02 & & $10^3$ & & 1.00 \\
{\bf\scshape pt} & & \textit{121} & & \textit{0.00} & & \color{violet} 1.0 & & \color{violet} 88 & & \color{violet} -0.04 & & 1.5 & & 9 & & -0.13 & & 205 & & 0.10 & & 90 & & \textbf{90} & & \textbf{-0.03} & & 2.5 & & 73 & & -0.05 & & 87 & & -0.04 & & 1.0 & & 87 & & -0.04 & & $10^3$ & & 1.00 \\
{\bf\scshape ru} & & \textit{156} & & \textit{0.00} & & \color{violet} 0.5 & & \color{violet} 70 & & \color{violet} -0.10 & & 5.0 & & 22 & & -0.16 & & 72 & & -0.10 & & 90 & & \textbf{90} & & \textbf{-0.08} & & 5.0 & & 21 & & -0.16 & & $10^3$ & & 1.00 & & 1.0 & & $10^3$ & & 1.00 & & $10^3$ & & 1.00 \\
 \hline
\end{tabular}%
}
\caption{  {\sc n}euro{\sc mst} encoder monitoring of local testing frames for resource-rich languages}
\label{table-monitoring-local-testing-frames-neuromst-resource-rich-languages}
\end{table} 

\begin{table}[htbp]
\centering
\resizebox{\textwidth}{!}{%
\tabcolsep=0.1cm
\begin{tabular}{rrrrrrrrrrrrrrrrrrrrrrrrrrrrrrrrrrrrrrrrrrrrrrr}
\hline
                 &  & \multicolumn{3}{c}{$c_{\wp[10^3]}$}                                              &                      & \multicolumn{5}{c}{\color{violet}$c_{\textit{coi}[\aleph,5,\alpha_{\aleph}]}$} &                      & \multicolumn{5}{c}{$c_{{\textit{gl}[\alpha]}}$}                                              &                      & \multicolumn{3}{c}{$c_{\textit{hnr}[5,5.0]}$}                                              &                      & \multicolumn{5}{c}{$c_{\textit{mne}[m]}$}                                                                        &                      & \multicolumn{5}{c}{$c_{\textit{og}[\alpha]}$}                                                                             &                      & \multicolumn{3}{c}{$c_{{\textit{p}[5,1.0]}}$}                                              &                      & \multicolumn{5}{c}{$c_{\textit{pq}[5,\alpha]}$}                                                                             &                      & \multicolumn{3}{c}{$c_{\textit{up}[5,5]}$}      \\                                         \cline{1-1} \cline{3-5} \cline{7-11} \cline{13-17} \cline{19-21} \cline{23-27} \cline{29-33} \cline{35-37} \cline{39-43} \cline{45-47}
                 &  & \multicolumn{1}{c}{$s_{c_{\wp[10^3]}}$} & & \multicolumn{1}{c}{$\Phi_{0.5}^{0.5}$} & \multicolumn{1}{c}{} & \multicolumn{1}{c}{\color{violet}$\alpha_{\aleph}$} & & \multicolumn{1}{c}{\color{violet} $s_{c_{\textit{coi}[\aleph,5,\alpha_{\aleph}]}}$} & & \multicolumn{1}{c}{\color{violet} $\Phi_{0.5}^{0.5}$} & \multicolumn{1}{c}{} & \multicolumn{1}{c}{$\alpha$} & \multicolumn{1}{c}{} & \multicolumn{1}{c}{$s_{c_{{\textit{gl}[\alpha]}}}$} & & \multicolumn{1}{c}{$\Phi_{0.5}^{0.5}$} & \multicolumn{1}{c}{} & \multicolumn{1}{c}{$s_{c_{\textit{hnr}[5,5.0]}}$} & & \multicolumn{1}{c}{$\Phi_{0.5}^{0.5}$} & \multicolumn{1}{c}{} & \multicolumn{1}{c}{$m$} & & \multicolumn{1}{c}{$s_{c_{\textit{mne}[m]}}$} & & \multicolumn{1}{c}{$\Phi_{0.5}^{0.5}$} & \multicolumn{1}{c}{} & \multicolumn{1}{c}{$\alpha$} & & \multicolumn{1}{c}{$s_{c_{\textit{og}[\alpha]}}$} & & \multicolumn{1}{c}{$\Phi_{0.5}^{0.5}$} & \multicolumn{1}{c}{} & \multicolumn{1}{c}{$s_{c_{{\textit{p}[5,1.0]}}}$} & & \multicolumn{1}{c}{$\Phi_{0.5}^{0.5}$} & \multicolumn{1}{c}{} & \multicolumn{1}{c}{$\alpha$} & & \multicolumn{1}{c}{$s_{c_{\textit{pq}[5,\alpha]}}$} & & \multicolumn{1}{c}{$\Phi_{0.5}^{0.5}$} & \multicolumn{1}{c}{} & \multicolumn{1}{c}{$s_{c_{\textit{up}[5,5]}}$} & & \multicolumn{1}{c}{$\Phi_{0.5}^{0.5}$} \\ \cline{1-1} \cline{3-3} \cline{5-5} \cline{7-7} \cline{9-9} \cline{11-11} \cline{13-13} \cline{15-15} \cline{17-17} \cline{19-19} \cline{21-21} \cline{23-23} \cline{25-25} \cline{27-27} \cline{29-29}  \cline{31-31} \cline{33-33} \cline{35-35} \cline{37-37} \cline{39-39} \cline{41-41} \cline{43-43} \cline{45-45} \cline{47-47}   
{\bf\scshape ca} & & \textit{277} & & \textit{0.00} & & \color{violet} 5.0 & & \color{violet} \textbf{277} & & \color{violet} \textbf{0.00} & & 1.0 & & 8 & & -0.37 & & 147 & & -0.18 & & 90 & & 90 & & -0.26 & & 3.0 & & 84 & & -0.27 & & 64 & & -0.29 & & 1.0 & & 64 & & -0.29 & & $10^3$ & & 1.00 \\
{\bf\scshape eu} & & \textit{762} & & \textit{0.00} & & \color{violet} 1.0 & & \color{violet} $10^3$ & & \color{violet} 0.32 & & 4.0 & & 13 & & -0.99 & & \textbf{114} & & \textbf{-0.87} & & 90 & & 90 & & -0.90 & & 0.5 & & 2 & & -0.99 & & $10^3$ & & 0.32 & & 1.0 & & $10^3$ & & 0.32 & & $10^3$ & & 0.32 \\
{\bf\scshape fa} & & \textit{342} & & \textit{0.00} & & \color{violet} 0.5 & & \color{violet} 94 & & \color{violet} -0.37 & & 1.0 & & 5 & & -0.50 & & \textbf{100} & & \textbf{-0.36} & & 90 & & 90 & & -0.38 & & 0.5 & & 5 & & -0.50 & & $10^3$ & & 1.00 & & 1.0 & & $10^3$ & & 1.00 & & $10^3$ & & 1.00 \\
{\bf\scshape gl} & & \textit{117} & & \textit{0.00} & & \color{violet} 5.0 & & \color{violet} \textbf{117} & & \color{violet} \textbf{0.00} & & 3.5 & & 15 & & -0.11 & & 98 & & -0.02 & & 90 & & 90 & & -0.03 & & 0.5 & & 2 & & -0.10 & & $10^3$ & & 1.00 & & 1.0 & & $10^3$ & & 1.00 & & 837 & & 0.82 \\
{\bf\scshape hi} & & \textit{24} & & \textit{0.00} & & \color{violet} 0.5 & & \color{violet} 28 & & \color{violet} 0.00 & & 1.0 & & 5 & & -0.02 & & 76 & & 0.05 & & 20 & & \textbf{20} & & \textbf{-0.00} & & 1.0 & & 17 & & -0.01 & & 35 & & 0.01 & & 1.0 & & 35 & & 0.01 & & $10^3$ & & 1.00 \\
{\bf\scshape ja} & & \textit{6} & & \textit{0.00} & & \color{violet} 0.5 & & \color{violet} \textbf{5} & & \color{violet} \textbf{0.00} & & 1.0 & & 3 & & -0.00 & & 44 & & 0.04 & & 10 & & 10 & & 0.00 & & 0.5 & & \textbf{5} & & \textbf{0.00} & & 12 & & 0.01 & & 1.0 & & 7 & & 0.00 & & $10^3$ & & 1.00 \\
{\bf\scshape sr} & & \textit{649} & & \textit{0.00} & & \color{violet} 0.7 & & \color{violet} \textbf{425} & & \color{violet} \textbf{-0.35} & & 1.0 & & 8 & & -1.00 & & 79 & & -0.89 & & 90 & & 90 & & -0.88 & & 5.0 & & 46 & & -0.94 & & $10^3$ & & 0.55 & & 1.0 & & $10^3$ & & 0.55 & & $10^3$ & & 0.55 \\
{\bf\scshape zh} & & \textit{31} & & \textit{0.00} & & \color{violet} 0.5 & & \color{violet} 24 & & \color{violet} -0.01 & & 5.0 & & 4 & & -0.01 & & 115 & & 0.09 & & 30 & & \textbf{30} & & \textbf{-0.00} & & 0.5 & & 3 & & -0.01 & & $10^3$ & & 1.00 & & 1.0 & & $10^3$ & & 1.00 & & 142 & & 0.12 \\ \hline
\end{tabular}%
}
\caption{{\sc n}euro{\sc mst} encoder monitoring of local testing frames for resource-poor languages}
\label{table-monitoring-local-testing-frames-neuromst-resource-poor-languages}
\end{table}

\paragraph{Comparison with state-of-the-art online indicators}

\begin{figure}[htbp]
\begin{center}
\begin{tabular}{cc}
\includegraphics[width=.97\textwidth]{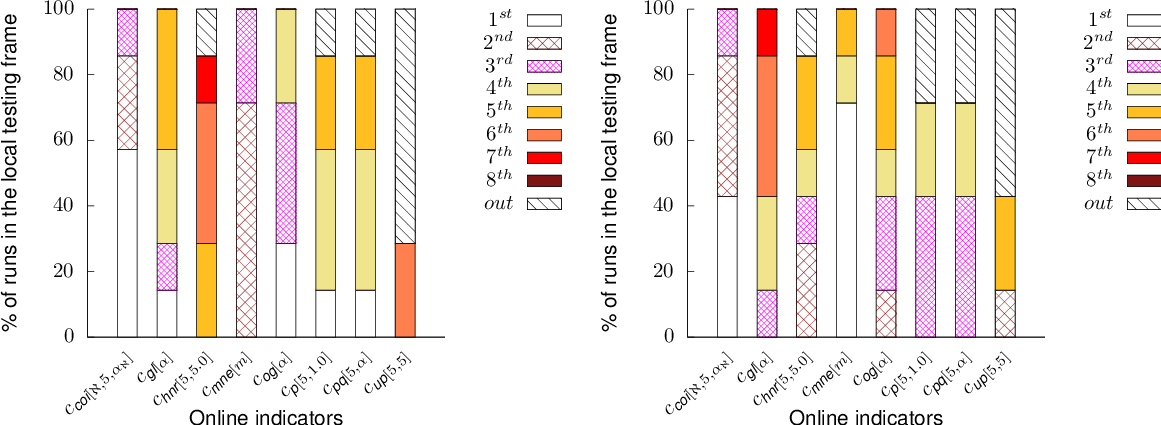}
\end{tabular}
\caption{{\sc b}i{\sc af} and {\sc n}euro{\sc mst}
            model ranking percentages across local testing
            frames for resource-rich languages}
\label{fig-percentages-ranking-positions-on-biaf-neuromst-resource-rich-languages}
\end{center}          
\end{figure}

\begin{figure}[htbp]
\begin{center}
\begin{tabular}{cc}
\includegraphics[width=.97\textwidth]{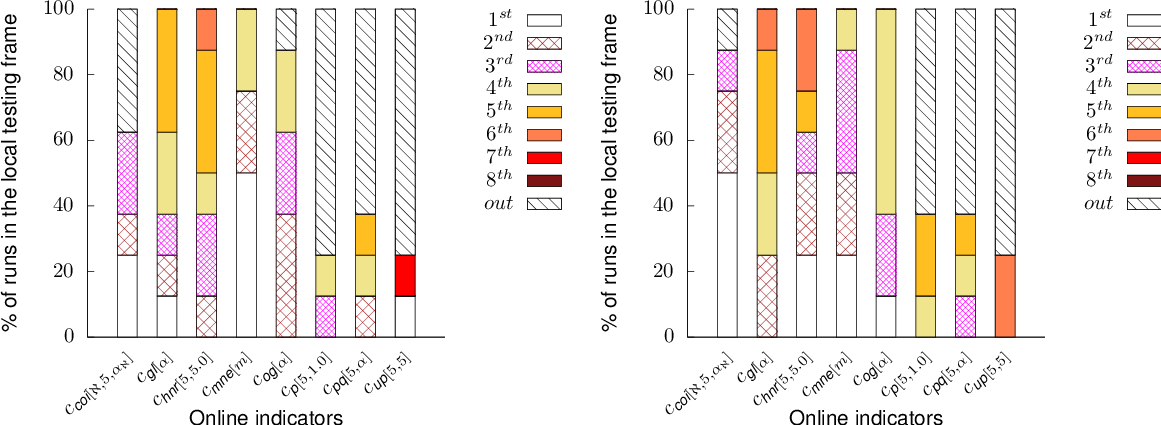}
\end{tabular}
\caption{{\sc b}i{\sc af} and {\sc n}euro{\sc mst}
            model ranking percentages across local testing
            frames for resource-poor languages}
\label{fig-percentages-ranking-positions-on-biaf-neuromst-resource-poor-languages}
\end{center}          
\end{figure}

In this case,
Figs.~\ref{fig-percentages-ranking-positions-on-biaf-neuromst-resource-rich-languages}
and~\ref{fig-percentages-ranking-positions-on-biaf-neuromst-resource-poor-languages}
show percentages for ranking positions across runs in local testing
frames, for the {\sc b}i{\sc af} (left-hand side) and {\sc n}euro{\sc
  mst} (right-hand side) models, when applied to resource-rich and
resource-poor languages, respectively.

Observable at first glance is the impact of both the neural
architecture and the size of the corpus. As expected, most indicators
perform significantly worse when applied to smaller datasets,
resorting to the primary rule to stop the learning process in some of
the runs studied, a circumstance that is easily identifiable from
Tables~\ref{table-monitoring-local-testing-frames-biaf-resource-rich-languages}
to~\ref{table-monitoring-local-testing-frames-neuromst-resource-poor-languages}
because epoch $s_{c_i}$ takes the value $10^3$ of the horizon $h$ set
for the oracle. All this corroborates the difficulty of categorizing
the state-of-the-art online indicators according to
performance~\citep{Lodwich:2009:ERP:1704175.1704224,Nguyen:2012:WSI:2437054.2437098},
while also highlighting the outstanding behavior of
$c_{\textit{coi}[\aleph,5,\alpha_\aleph]}$ whatever the scenario
considered.

More specifically, for large enough corpora, the {\sc b}i{\sc af}
models monitored by $c_{\textit{coi}[\aleph,5,\alpha_\aleph]}$ achieve
the top \nemesid{} scores in most local testing frames ({\sc en},
{\sc es}, {\sc pl} and {\sc pt}) reflected in
Table~\ref{table-monitoring-local-testing-frames-biaf-resource-rich-languages}
but three, while it ranked second ({\sc fr} and {\sc ru}) and
third ({\sc de}) best in the rest. Regarding the {\sc n}euro{\sc
  mst} architecture, the performance is similar, achieving the top
\nemesid{} scores in three local testing frames ({\sc en}, {\sc es}
and {\sc fr}) included in
Table~\ref{table-monitoring-local-testing-frames-neuromst-resource-rich-languages},
while it ranks second ({\sc de}, {\sc pl} and {\sc pt}) and third
({\sc ru}) best in the rest.

\paragraph{Comparison with the oracle}

Focusing on large enough corpora and using a {\sc b}i{\sc af} encoder,
$c_{\textit{coi}[\aleph,5,\alpha_\aleph]}$ is as accurate as the
oracle in two of the local testing frames ({\sc en} and {\sc es}),
although the difference from the baseline never exceeds two epochs,
except for the corpus {\sc de}, as shown in
Table~\ref{table-monitoring-local-testing-frames-biaf-resource-rich-languages}. Specifically,
in that latter case $c_{\textit{coi}[\aleph,5,\alpha_\aleph]}$ stops
the learning process far in advance (118 epochs) of the oracle, but
with an acceptable cost/benefit trade-off, as is proven by its
\nemesid{} score (-0.15). The numbers worsen slightly for the {\sc
  n}euro{\sc mst} models which, although never matching the diagnosis
of the baseline, are a few iterations away from the oracle in two
cases ({\sc de} and {\sc es}) and provide good \nemesid{} scores (from
-0.10 up to 0.02) in all the other ones ({\sc en}, {\sc fr}, {\sc pl},
{\sc pt} and {\sc ru}), as can be seen in
Table~\ref{table-monitoring-local-testing-frames-neuromst-resource-rich-languages}. This
translates into a reduction (from 1 up to 86 epochs) of model
generation costs except in one case ({\sc en}), always in
combination with a very good cost/benefit trade-off.

As for small corpora and excluding out-of-range runs, our proposal
matches the oracle only once when using {\sc b}i{\sc af} models ({\sc
  zh}), although the difference from the baseline is always negative
and does not exceed a few epochs (from 1 up to 9) in other three
cases ({\sc ca}, {\sc gl} and {\sc hi}), which means a slight
reduction in model generation costs, as shown in
Table~\ref{table-monitoring-local-testing-frames-biaf-resource-poor-languages}.
For the remaining run ({\sc sr}),
$c_{\textit{coi}[\aleph,5,\alpha_\aleph]}$ stops the learning process
far in advance (192 epochs) of the oracle, but with an acceptable
cost/benefit trade-off, as shown in
Table~\ref{table-monitoring-local-testing-frames-biaf-resource-poor-languages}
through its \nemesid{} score of -0.27. These numbers improve slightly
for {\sc n}euro{\sc mst} models, which match the baseline on two local
testing frames ({\sc ca} and {\sc gl}) and barely differ in a few
epochs (from 1 up to 6 epochs) in three other ones ({\sc hi},
{\sc ja} and {\sc zh}). For the remaining two runs ({\sc fa} and {\sc
  sr}), $c_{\textit{coi}[\aleph,5,\alpha_\aleph]}$ stops the
learning process far in advance (248 and 224 epochs) of the
oracle, with an acceptable cost/benefit trade-off, as can be seen in
Table~\ref{table-monitoring-local-testing-frames-neuromst-resource-poor-languages}
through their \nemesid{} scores (-0.37 and -0.35).

\paragraph{Overview}

Overall, we find that the performance of our proposal not only
improves on the state-of-the-art, but also comes very close to the
oracle, regardless of the architecture applied by the learner or the
type of corpus under consideration. All this in operational conditions
that are not particularly advantageous for it as we sought the best
fit for the rest of the online indicators tested, which is not the
case for $c_{\textit{coi}[\aleph,k,\alpha_\aleph]}$, whose parameters
$\aleph$ and $k$ were set for the occasion without an exhaustive prior
tuning process. In particular, the length $k$ of of the training
strips, used in our case to delimit the sequence of epochs to be
correlated, has been set to the value recommended for the rest of the
indicators that apply this type of structure.

\subsubsection{The qualitative study}

Having confirmed the accuracy of our proposal as tested on a
particular configuration, $c_{\textit{coi}[\aleph,k,\alpha_\aleph]}$
with $k=5$, we now study its stability, to check whether the
good results observed show uniform behavior independent of the
resources used, and can also be extrapolated without major variations
to other settings. This would allow its use to be simplified,
as it would no longer require a cumbersome prior tuning phase.

\paragraph{Stability with regard to the resources}

\begin{figure}[htbp]
\begin{center}
\begin{tabular}{cc}
\includegraphics[width=.97\textwidth]{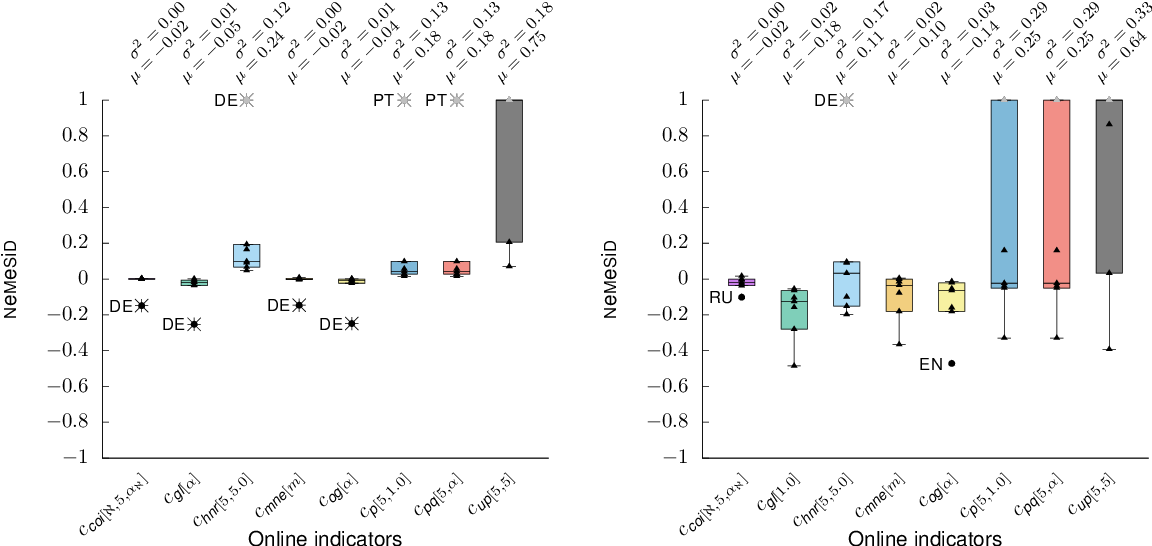}
\end{tabular}
\caption{\nemesid{} {\sc b}i{\sc af} and {\sc n}euro{\sc mst} model box plots across local testing frames for resource-rich languages}
\label{fig-nemesid-box-plots-on-biaf-neuromst-resource-rich-languages}
\end{center}          
\end{figure}

Concerning this,
Fig.~\ref{fig-nemesid-box-plots-on-biaf-neuromst-resource-rich-languages}
shows that the \nemesid{} scores for
$c_{\textit{coi}[\aleph,5,\alpha_\aleph]}$ for the resource-rich
languages are the most stable among all the indicators evaluated,
independent of the learning architecture used. More to the point, it
virtually matches the optimal performance when using a {\sc b}i{\sc
  af} model, excluding the run corresponding to {\sc de}, which turns
out to be an extreme or even an out-of-range run for most
indicators. In fact, all the stop conditions exhibit such values,
along with --except for $c_{\textit{coi}[\aleph,5,\alpha_\aleph]}$
and $c_{\textit{mne}[m]}$-- a varying level of asymmetry both in
terms of interquartile range and whisker length, revealing an
increasing dispersion and variability of the \nemesid{} values.

As for {\sc n}euro{\sc mst}, although slightly different to those of the
oracle, the values for $c_{\textit{coi}[\aleph,5,\alpha_\aleph]}$ are
still excellent, and the best overall. In greater detail, it shows a
mean close to the optimum, with symmetry at the interquartile range
and a slight deviation in the whiskers, which show the smallest range and
the shortest length respectively. The result is the practical
absence of biases in the distribution of \nemesid{} values, with
little variability and a high concentration around those of the
oracle. In contrast, the impact of the neural architecture is high on
the rest of the indicators. Thus, both the size and the asymmetry of
interquartile boxes and whiskers increase, while the mean moves away
from that of the oracle. Therefore, accuracy and stability degrade
significantly with out-of-range runs for half of the indicators,
although most of the extreme values have dissapeared with respect to the
{\sc b}i{\sc af} encoders.



\begin{figure}[htbp]
\begin{center}
\begin{tabular}{cc}
\includegraphics[width=.97\textwidth]{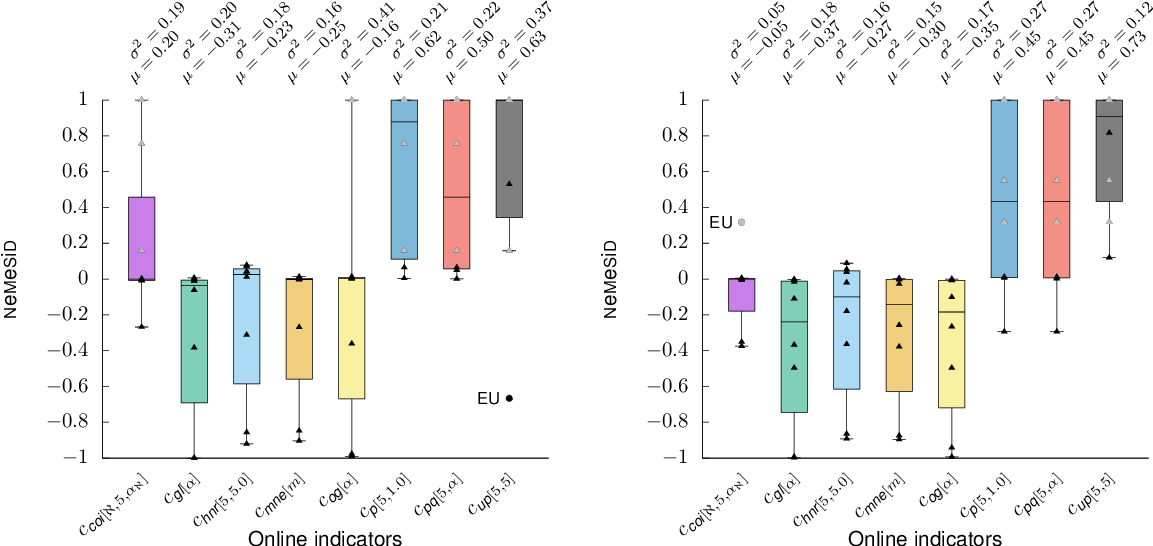}    
\end{tabular}
\caption{\nemesid{} {\sc
       b}i{\sc af} and {\sc n}euro{\sc mst} model box plots across local testing frames for resource-poor
       languages}
\label{fig-nemesid-box-plots-on-biaf-neuromst-resource-poor-languages}
\end{center}          
\end{figure}

Regarding the \nemesid{} scores for the resource-poor languages,
Fig.~\ref{fig-nemesid-box-plots-on-biaf-neuromst-resource-poor-languages}
shows an increased interquartile range and whisker length for most of
the indicators with a worsening of the asymmetry in both cases, which
implies the presence of biases. In this respect,
$c_{\textit{coi}[\aleph,5,\alpha_\aleph]}$ is associated with the most
contained distribution, while the symmetry is suffering from the
impact of the lack of learning resources, although to a lesser extent
than the rest of the indicators studied. The only exception to this
generalized behavior is (resp. are) $c_{{\textit{pq}[5,\alpha]}}$
(resp. $c_{\textit{p}[5, 1.0]}$ and $c_{{\textit{pq}[5,\alpha]}}$)
when using a {\sc b}i{\sc af} (resp. {\sc n}euro{\sc mst})
architecture, but their interquartile ranges and whisker lengths are
the highest, which implies a significant dispersion and variability in
the performance of the various runs. In contrast to what happened with
the sufficiently large corpora, most outliers and extreme values have
disappeared here, although the number of out-of-range runs has
increased, concentrated mainly among the same indicators as for the
large ones.

\paragraph{Stability with regard to the setting}

Regarding the stability of $c_{\textit{coi}[\aleph,k,\alpha_\aleph]}$
for various settings, we focus on the length $k$ of the training strip
since the tuning of the threshold $\alpha_\aleph$ for each local
testing frame had already been described, and we assume that the
impact of the concrete correlation coefficient $\aleph$ used is
comparatively minor in regulating the learning process. Bearing in
mind that a correct strip must have a length of at least 2, we compare
the \nemesid{} values obtained for $k=2, 5$ and $8$, ensuring a
symmetric analysis. Once again taking cross-validation as the canary
function $c$, we introduce the collection $\Upsilon_c^{\textit{coi}}$
of online indicators to be considered as:
\begin{equation}
\label{eq-collection-online-indicators-COI-testing-frame}
  \Upsilon_c^{\textit{coi}} := \{ c_{\textit{coi}[\aleph,2,\alpha_\aleph]}, c_{\textit{coi}[\aleph,5,\alpha_\aleph]}, c_{\textit{coi}[\aleph,8,\alpha_\aleph]}\}
\end{equation}
\noindent together with the neural architectures ({\sc b}i{\sc af} and
{\sc n}euro{\sc mst}) and the {\sc ud} treebank collections
($R$ and $P$) as already identified. We then define
the  local testing frame families as:
\begin{equation}
\label{eq-families-local-testing-frames-2}
  \{ \mathcal{L}_{R,d}^{\Upsilon_c^{\textit{coi}}}\}_{d \in {\mathcal D}} \mbox{ (resp. }
 \{ \mathcal{L}_{P,d}^{\Upsilon_c^{\textit{coi}}}\}_{d \in {\mathcal D}}\mbox{)}, \mbox{ with } {\mathcal D} := \{\mbox{{\sc b}i{\sc af}}, \mbox{{\sc n}euro{\sc mst}}\}
\end{equation}
\noindent one \textit{per} combination of corpus in $R$
(resp.  $P$) and encoder $d \in {\mathcal D}$.  Those
families, together with the \nemesid{} quality metric make up our new
experimental environment.



\begin{table}[htbp]
\centering
\resizebox{\textwidth}{!}{%
\tabcolsep=0.1cm
\begin{tabular}{rrrrrrrrrrrrrrrrrrrrrrrrrrrrrrrrrrrrrrr}
\cline{1-19} \cline{21-39}
\multicolumn{1}{c}{} & \multicolumn{1}{c}{} & \multicolumn{5}{c}{$c_{\textit{coi}[\aleph,2,\alpha_{\aleph}]}$}                                                                                                                                    & \multicolumn{1}{c}{} & \multicolumn{5}{c}{\color{violet} $c_{\textit{coi}[\aleph,5,\alpha_{\aleph}]}$}                                                                                                                                    & \multicolumn{1}{c}{} & \multicolumn{5}{c}{$c_{\textit{coi}[\aleph,8,\alpha_{\aleph}]}$}                                                                                                                                    &                      & \multicolumn{1}{c}{} & \multicolumn{1}{l}{} & \multicolumn{5}{c}{$c_{\textit{coi}[\aleph,2,\alpha_{\aleph}]}$}                                                                                                                                    & \multicolumn{1}{c}{} & \multicolumn{5}{c}{$c_{\color{violet} \textit{coi}[\aleph,5,\alpha_{\aleph}]}$}                                                                                                                                    & \multicolumn{1}{c}{} & \multicolumn{5}{c}{$c_{\textit{coi}[\aleph,8,\alpha_{\aleph}]}$}                                                                                                                                    \\ \cline{1-1} \cline{3-7} \cline{9-13} \cline{15-19} \cline{21-21} \cline{23-27} \cline{29-33} \cline{35-39} 
\multicolumn{1}{c}{} & \multicolumn{1}{c}{} & \multicolumn{1}{c}{$\alpha_{\aleph}$} & \multicolumn{1}{c}{} & \multicolumn{1}{c}{$s_{c_{\textit{coi}[\aleph,2,\alpha_{\aleph}]}}$} & \multicolumn{1}{c}{} & \multicolumn{1}{c}{$\Phi_{0.5}^{0.5}$} & \multicolumn{1}{c}{} & \multicolumn{1}{c}{\color{violet} $\alpha_{\aleph}$} & \multicolumn{1}{c}{} & \multicolumn{1}{c}{\color{violet} $s_{c_{\textit{coi}[\aleph,5,\alpha_{\aleph}]}}$} & \multicolumn{1}{c}{} & \multicolumn{1}{c}{\color{violet} $\Phi_{0.5}^{0.5}$} & \multicolumn{1}{c}{} & \multicolumn{1}{c}{$\alpha_{\aleph}$} & \multicolumn{1}{c}{} & \multicolumn{1}{c}{$s_{c_{\textit{coi}[\aleph,8,\alpha_{\aleph}]}}$} & \multicolumn{1}{c}{} & \multicolumn{1}{c}{$\Phi_{0.5}^{0.5}$} &                      & \multicolumn{1}{c}{} & \multicolumn{1}{l}{} & \multicolumn{1}{c}{$\alpha_{\aleph}$} & \multicolumn{1}{l}{} & \multicolumn{1}{c}{$s_{c_{\textit{coi}[\aleph,2,\alpha_{\aleph}]}}$} & \multicolumn{1}{l}{} & \multicolumn{1}{c}{$\Phi_{0.5}^{0.5}$} & \multicolumn{1}{l}{} & \multicolumn{1}{c}{\color{violet} $\alpha_{\aleph}$} & \multicolumn{1}{l}{} & \multicolumn{1}{c}{\color{violet} $s_{c_{\textit{coi}[\aleph,5,\alpha_{\aleph}]}}$} & \multicolumn{1}{l}{} & \multicolumn{1}{c}{\color{violet} $\Phi_{0.5}^{0.5}$} & \multicolumn{1}{l}{} & \multicolumn{1}{c}{$\alpha_{\aleph}$} & \multicolumn{1}{l}{} & \multicolumn{1}{c}{$s_{c_{\textit{coi}[\aleph,8,\alpha_{\aleph}]}}$} & \multicolumn{1}{l}{} & \multicolumn{1}{c}{$\Phi_{0.5}^{0.5}$} \\ \cline{1-1} \cline{3-3} \cline{5-5} \cline{7-7} \cline{9-9} \cline{11-11} \cline{13-13} \cline{15-15} \cline{17-17} \cline{19-19} \cline{21-21} \cline{23-23} \cline{25-25} \cline{27-27} \cline{29-29} \cline{31-31} \cline{33-33} \cline{35-35} \cline{37-37} \cline{39-39} 
{\bf\scshape de} & & 0.5 & & \textbf{196} & & \textbf{-0.01} & & \color{violet} 0.5 & & \color{violet} 88 & & \color{violet} -0.15 & & 0.5 & & 92 & & -0.14 & & {\bf\scshape ca} & & 0.5 & & 40 & & -0.03 & & \color{violet} 0.7 & & \color{violet} 55 & & \color{violet} -0.01 & & 0.7 & & \textbf{58} & & \textbf{-0.01} \\
{\bf\scshape en} & & 0.5 & & \textbf{41} & & \textbf{0.00} & & \color{violet} 0.5 & & \color{violet} \textbf{41} & & \color{violet} \textbf{0.00} & & 0.5 & & 46 & & 0.01 & & {\bf\scshape eu} & & 0.5 & & $10^3$ & & 0.76 & & \color{violet} 1.0 & & \color{violet} $10^3$ & & \color{violet} 0.76 & & 1.0 & & $10^3$ & & 0.76 \\
{\bf\scshape es} & & 0.5 & & \textbf{42} & & \textbf{0.00} & & \color{violet} 0.5 & & \color{violet} \textbf{42} & & \color{violet} \textbf{0.00} & & 0.5 & & 43 & & 0.00 & & {\bf\scshape fa} & & 1.0 & & $10^3$ & & 0.16 & & \color{violet} 1.0 & & \color{violet} $10^3$ & & \color{violet} 0.16 & & 1.0 & & $10^3$ & & 0.16 \\
{\bf\scshape fr} & & 0.5 & & \textbf{4} & & \textbf{0.00} & & \color{violet} 0.5 & & \color{violet} 5 & & \color{violet} 0.00 & & 0.5 & & 8 & & 0.01 & & {\bf\scshape gl} & & 0.5 & & \textbf{19} & & \textbf{-0.00} & & \color{violet} 0.5 & & \color{violet} \textbf{19} & & \color{violet} \textbf{-0.00} & & 0.5 & & 17 & & -0.00 \\
{\bf\scshape pl} & & 0.5 & & 20 & & -0.00 & & \color{violet} 0.5 & & \color{violet} 21 & & \color{violet} -0.00 & & 0.5 & & \textbf{22} & & \textbf{0.00} & & {\bf\scshape hi} & & 0.5 & & 10 & & -0.01 & & \color{violet} 0.5 & & \color{violet} 13 & & \color{violet} -0.00 & & 0.5 & & \textbf{16} & & \textbf{-0.00} \\
{\bf\scshape pt} & & 0.5 & & \textbf{27} & & \textbf{0.00} & & \color{violet} 0.7 & & \color{violet} 29 & & \color{violet} 0.00 & & 0.5 & & 23 & & -0.00 & & {\bf\scshape ja} & & 1.0 & & $10^3$ & & 1.00 & & \color{violet} 1.0 & & \color{violet} $10^3$ & & \color{violet} 1.00 & & 1.0 & & $10^3$ & & 1.00 \\
{\bf\scshape ru} & & 0.5 & & \textbf{14} & & \textbf{0.00} & & \color{violet} 0.5 & & \color{violet} 16 & & \color{violet} 0.00 & & 0.5 & & 8 & & -0.01 & & {\bf\scshape sr} & & 0.5 & & \textbf{264} & & \textbf{-0.03} & & \color{violet} 0.5 & & \color{violet} 92 & & \color{violet} -0.27 & & 0.5 & & 90 & & -0.27 \\
 & &  & &  & &  & &  & &  & &  & &  & &  & &  & & {\bf\scshape zh} & & 0.5 & & \textbf{5} & & \textbf{0.00} & & \color{violet} 0.5 & & \color{violet} \textbf{5} & & \color{violet} \textbf{0.00} & & 0.5 & & 8 & & 0.01 \\                                   \cline{1-19} \cline{21-39} 
\end{tabular}%
}
\caption{{\sc b}i{\sc af} encoder monitoring of local testing frames with $c_{\textit{coi}[\aleph,k,\alpha_\aleph]}$ for resource-rich and resource-poor languages}
\label{table-monitoring-COI-local-testing-frames-biaf}
\end{table}

\begin{table}[htbp]
\centering
\resizebox{\textwidth}{!}{%
\tabcolsep=0.1cm
\begin{tabular}{rrrrrrrrrrrrrrrrrrrrrrrrrrrrrrrrrrrrrrr}
\cline{1-19} \cline{21-39}
\multicolumn{1}{c}{} & \multicolumn{1}{c}{} & \multicolumn{5}{c}{$c_{\textit{coi}[\aleph,2,\alpha_{\aleph}]}$}                                                                                                                                    & \multicolumn{1}{c}{} & \multicolumn{5}{c}{\color{violet} $c_{\textit{coi}[\aleph,5,\alpha_{\aleph}]}$}                                                                                                                                    & \multicolumn{1}{c}{} & \multicolumn{5}{c}{$c_{\textit{coi}[\aleph,8,\alpha_{\aleph}]}$}                                                                                                                                    &                      & \multicolumn{1}{c}{} & \multicolumn{1}{l}{} & \multicolumn{5}{c}{$c_{\textit{coi}[\aleph,2,\alpha_{\aleph}]}$}                                                                                                                                    & \multicolumn{1}{c}{} & \multicolumn{5}{c}{$c_{\color{violet} \textit{coi}[\aleph,5,\alpha_{\aleph}]}$}                                                                                                                                    & \multicolumn{1}{c}{} & \multicolumn{5}{c}{$c_{\textit{coi}[\aleph,8,\alpha_{\aleph}]}$}                                                                                                                                    \\ \cline{1-1} \cline{3-7} \cline{9-13} \cline{15-19} \cline{21-21} \cline{23-27} \cline{29-33} \cline{35-39} 
\multicolumn{1}{c}{} & \multicolumn{1}{c}{} & \multicolumn{1}{c}{$\alpha_{\aleph}$} & \multicolumn{1}{c}{} & \multicolumn{1}{c}{$s_{c_{\textit{coi}[\aleph,2,\alpha_{\aleph}]}}$} & \multicolumn{1}{c}{} & \multicolumn{1}{c}{$\Phi_{0.5}^{0.5}$} & \multicolumn{1}{c}{} & \multicolumn{1}{c}{\color{violet} $\alpha_{\aleph}$} & \multicolumn{1}{c}{} & \multicolumn{1}{c}{\color{violet} $s_{c_{\textit{coi}[\aleph,5,\alpha_{\aleph}]}}$} & \multicolumn{1}{c}{} & \multicolumn{1}{c}{\color{violet} $\Phi_{0.5}^{0.5}$} & \multicolumn{1}{c}{} & \multicolumn{1}{c}{$\alpha_{\aleph}$} & \multicolumn{1}{c}{} & \multicolumn{1}{c}{$s_{c_{\textit{coi}[\aleph,8,\alpha_{\aleph}]}}$} & \multicolumn{1}{c}{} & \multicolumn{1}{c}{$\Phi_{0.5}^{0.5}$} &                      & \multicolumn{1}{c}{} & \multicolumn{1}{l}{} & \multicolumn{1}{c}{$\alpha_{\aleph}$} & \multicolumn{1}{l}{} & \multicolumn{1}{c}{$s_{c_{\textit{coi}[\aleph,2,\alpha_{\aleph}]}}$} & \multicolumn{1}{l}{} & \multicolumn{1}{c}{$\Phi_{0.5}^{0.5}$} & \multicolumn{1}{l}{} & \multicolumn{1}{c}{\color{violet} $\alpha_{\aleph}$} & \multicolumn{1}{l}{} & \multicolumn{1}{c}{\color{violet} $s_{c_{\textit{coi}[\aleph,5,\alpha_{\aleph}]}}$} & \multicolumn{1}{l}{} & \multicolumn{1}{c}{\color{violet} $\Phi_{0.5}^{0.5}$} & \multicolumn{1}{l}{} & \multicolumn{1}{c}{$\alpha_{\aleph}$} & \multicolumn{1}{l}{} & \multicolumn{1}{c}{$s_{c_{\textit{coi}[\aleph,8,\alpha_{\aleph}]}}$} & \multicolumn{1}{l}{} & \multicolumn{1}{c}{$\Phi_{0.5}^{0.5}$} \\ \cline{1-1} \cline{3-3} \cline{5-5} \cline{7-7} \cline{9-9} \cline{11-11} \cline{13-13} \cline{15-15} \cline{17-17} \cline{19-19} \cline{21-21} \cline{23-23} \cline{25-25} \cline{27-27} \cline{29-29} \cline{31-31} \cline{33-33} \cline{35-35} \cline{37-37} \cline{39-39} 
{\bf\scshape de} & & 0.5 & & \textbf{71} & & \textbf{0.01} & & \color{violet} 0.5 & & \color{violet} 61 & & \color{violet} -0.01 & & 0.5 & & 60 & & -0.01 & & {\bf\scshape ca} & & 0.5 & & 261 & & -0.02 & & \color{violet} 0.5 & & \color{violet} \textbf{277} & & \color{violet} \textbf{0.00} & & 1.0 & & 283 & & 0.01 \\
{\bf\scshape en} & & 0.5 & & \textbf{343} & & \textbf{0.01} & & \color{violet} 0.5 & & \color{violet} 345 & & \color{violet} 0.02 & & 0.7 & & 319 & & -0.02 & & {\bf\scshape eu} & & 0.5 & & $10^3$ & & 0.32 & & \color{violet} 0.5 & & \color{violet} $10^3$ & & \color{violet} 0.32 & & 0.5 & & $10^3$ & & 0.32 \\
{\bf\scshape es} & & 0.5 & & \textbf{71} & & \textbf{0.00} & & \color{violet} 0.5 & & \color{violet} 70 & & \color{violet} -0.00 & & 0.6 & & 67 & & -0.00 & & {\bf\scshape fa} & & 0.5 & & 91 & & -0.38 & & \color{violet} 0.5 & & \color{violet} 94 & & \color{violet} -0.37 & & 1.0 & & \textbf{343} & & \textbf{0.00} \\
{\bf\scshape fr} & & 0.5 & & \textbf{208} & & \textbf{-0.03} & & \color{violet} 0.5 & & \color{violet} \textbf{208} & & \color{violet} \textbf{-0.03} & & 0.5 & & 261 & & 0.04 & & {\bf\scshape gl} & & 0.5 & & 114 & & -0.00 & & \color{violet} 0.5 & & \color{violet} \textbf{117} & & \color{violet} \textbf{0.00} & & 0.5 & & 18 & & -0.11 \\
{\bf\scshape pl} & & 0.5 & & 85 & & -0.02 & & \color{violet} 0.5 & & \color{violet} 85 & & \color{violet} -0.02 & & 0.8 & & \textbf{103} & & \textbf{0.00} & & {\bf\scshape hi} & & 0.5 & & 17 & & -0.01 & & \color{violet} 0.5 & & \color{violet} \textbf{28} & & \color{violet} \textbf{0.00} & & 0.8 & & 17 & & -0.01 \\
{\bf\scshape pt} & & 0.5 & & 80 & & -0.05 & & \color{violet} 0.5 & & \color{violet} \textbf{88} & & \color{violet} \textbf{-0.04} & & 0.5 & & \textbf{88} & & \textbf{-0.04} & & {\bf\scshape ja} & & 0.5 & & \textbf{5} & & \textbf{-0.00} & & \color{violet} 0.5 & & \color{violet} \textbf{5} & & \color{violet} \textbf{-0.00} & & 0.5 & & 8 & & 0.00 \\
{\bf\scshape ru} & & 0.5 & & \textbf{90} & & \textbf{-0.08} & & \color{violet} 0.5 & & \color{violet} 70 & & \color{violet} -0.10 & & 0.5 & & 60 & & -0.11 & & {\bf\scshape sr} & & 0.5 & & $10^3$ & & 0.55 & & \color{violet} 0.7 & & \color{violet} 425 & & \color{violet} -0.35 & & 1.0 & & \textbf{439} & & \textbf{-0.33} \\
 & &  & &  & &  & &  & &  & &  & &  & &  & &  & & {\bf\scshape zh} & & 0.5 & & 3 & & -0.01 & & \color{violet} 0.5 & & \color{violet} \textbf{24} & & \color{violet} \textbf{-0.01} & & 0.5 & & 20 & & -0.01 \\ \cline{1-19} \cline{21-39} 
\end{tabular}%
}
\caption{{\sc n}euro{\sc mst} encoder monitoring of local testing frames with $c_{\textit{coi}[\aleph,k,\alpha_\aleph]}$ for resource-rich and resource-poor languages}
\label{table-monitoring-COI-local-testing-frames-neuromst}
\end{table}

\begin{figure}[htbp]
\begin{center}
\begin{tabular}{cc}
\includegraphics[width=.97\textwidth]{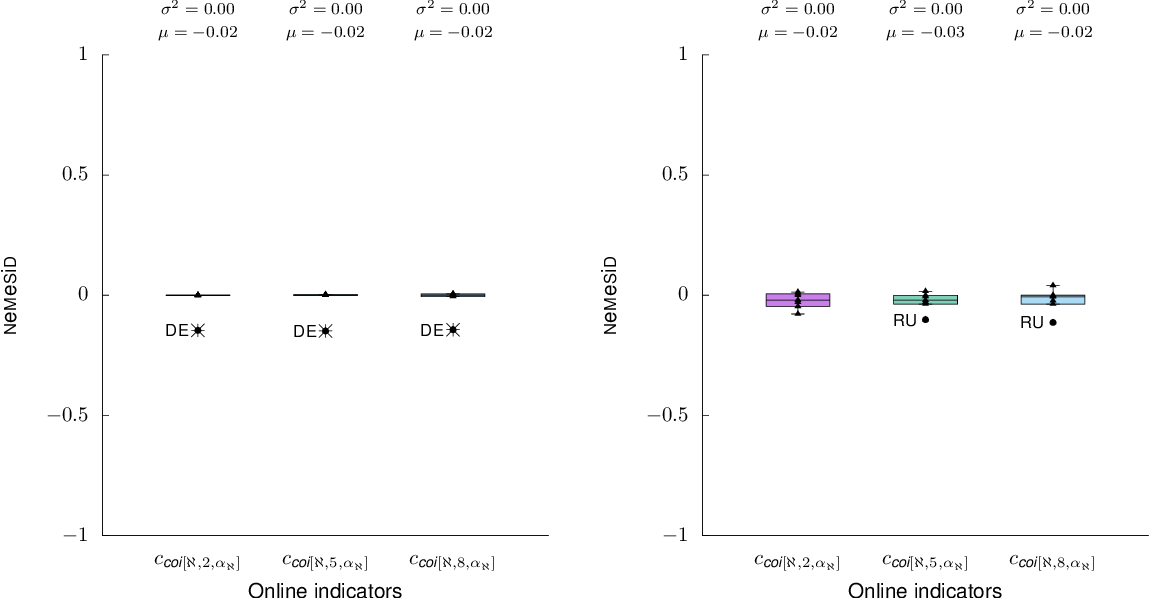}
\end{tabular}
\caption{\nemesid{} {\sc
       b}i{\sc af} and {\sc n}euro{\sc mst} model   box plots across {\sc coi}-based local testing frames for resource-rich
       languages}
\label{fig-COI-nemesid-box-plots-on-biaf-neuromst-resource-rich-languages}
\end{center}          
\end{figure}

\begin{figure}[htbp]
\begin{center}
\begin{tabular}{cc}
\includegraphics[width=.97\textwidth]{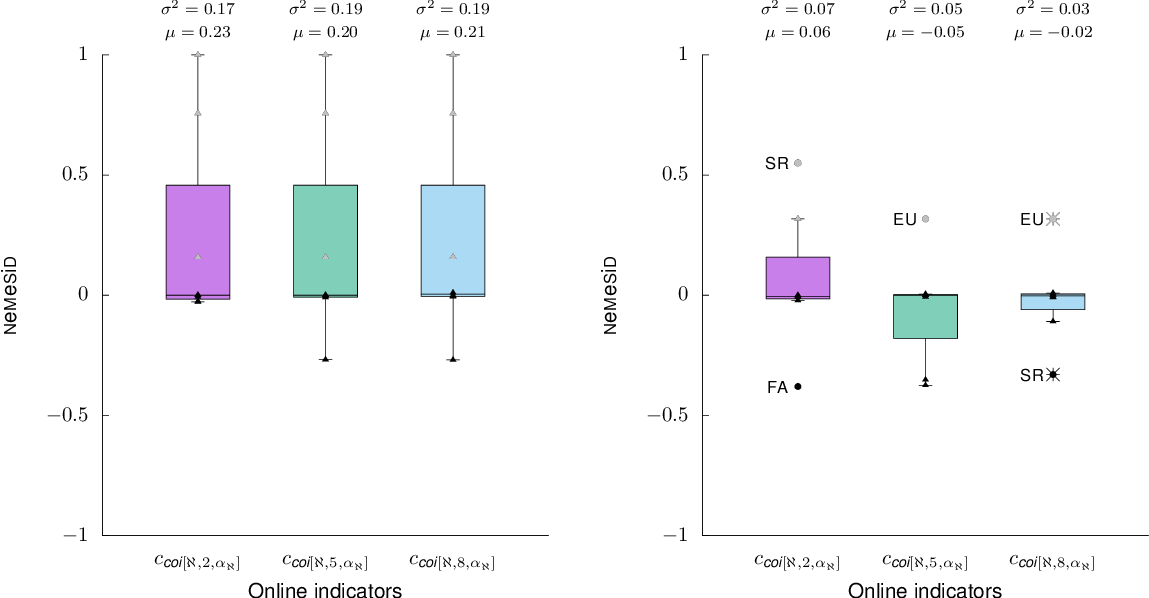}
\end{tabular}
\caption{\nemesid{} {\sc
       b}i{\sc af} and {\sc n}euro{\sc mst} model box plots across {\sc coi}-based local testing frames for resource-poor
       languages}
\label{fig-COI-nemesid-box-plots-on-biaf-neuromst-resource-poor-languages}
\end{center}          
\end{figure}

The rows in
Tables~\ref{table-monitoring-COI-local-testing-frames-biaf}
and~\ref{table-monitoring-COI-local-testing-frames-neuromst} reflect
the results for the local testing frames in
$\{\mathcal{L}_{R,d}^{\Upsilon_c^{\textit{coi}}},
\mathcal{L}_{P,d}^{\Upsilon_c^{\textit{coi}}}\}_{d \in {\mathcal D}}$,
and their entries show the runs associated with indicators in the
collection $\Upsilon_c^{\textit{coi}}$ following the same pattern
applied in previous tests. Meanwhile,
Figs.~\ref{fig-COI-nemesid-box-plots-on-biaf-neuromst-resource-rich-languages}
and~\ref{fig-COI-nemesid-box-plots-on-biaf-neuromst-resource-poor-languages}
summarize the distributions of \nemesid{} values associated with those
runs, differentiating between the {\sc b}i{\sc af} (left-hand side)
and the {\sc n}euro{\sc mst} (right-hand side) encodings for the
resource-rich and resource-poor languages, respectively.


Our proposal shows rock-solid stability for resource-rich languages,
with \nemesid{} values ranging over the interval $[-0.15, 0.01]$
(resp. $[-0.11, 0.04]$) for the {\sc b}i{\sc af} (resp. {\sc
  n}euro{\sc mst}) architecture, independent of the value $k \in
\{2,5,8\}$ considered for the length of the training strip, as
reflected in the left-hand-side of
Table~\ref{table-monitoring-COI-local-testing-frames-biaf}
(resp. Table~\ref{table-monitoring-COI-local-testing-frames-neuromst}),
although the best results are associated with $k=2$. Thus, the
interquartile range, whisker length and symmetry have no, or only minor,
fluctuations. Only the {\sc n}euro{\sc mst} models suffer a slight
degradation in the symmetry for the highest values of $k$, which
involves the whisker lengths for $k=5$, and both interquartile range
and whisker lengths for $k=8$. The presence of one extreme value for
the {\sc b}i{\sc af} encoders can be considered to be
irrelevant\footnote{Extreme values here reflect scores that deserve
  this qualification simply because the indicator is absolutely
  accurate for the rest of the corpora, as evidenced by the variance.}
and is once again associated with the {\sc de} corpus, as was the case for
most of the state-of-the-art indicators already studied. This suggests
that the origin of the phenomenon may lie to some extent in the nature
of the training database. Along the same lines, we can also consider the
characterization as an outlier of the run corresponding to the corpus
{\sc ru} when using {\sc n}euro{\sc mst}, for the cases $k=5$ and $k=8$,
as not very relevant.

Stability parameters worsen slightly for smaller corpora, if we
exclude out-of-range runs, with \nemesid{} values ranging over the
interval $[-0.27,0.01]$ (resp. $[-0.38,0.01]$) for {\sc b}i{\sc af}
(resp. {\sc n}euro{\sc mst}) models, independent of the value $k$
considered. On the other hand, and in contrast to what happened for
the larger corpora, the impact of architecture seems to be decisive
here. As for the distribution parameters, the impact of out-of-range
runs in the {\sc b}i{\sc af} architecture is visible, to which must be
added those associated with outliers and extreme values when using
{\sc n}euro{\sc mst} encoders. For the {\sc b}i{\sc af} case, this
impact is practically the same for all the various training strip
lengths considered, in terms of interquartile box, whiskers and
symmetry considerations. The only exception to this uniform behavior
is the length of the lower whisker for
$c_{{\textit{coi}[\aleph,2,\alpha\aleph]}}$, which is more closely
matched to the mean. In contrast, {\sc n}euro{\sc mst} models show a
more contained dispersion at all levels, but are also less uniform,
with the best distribution parameters associated with
$c_{{\textit{coi}[\aleph,8,\alpha\aleph]}}$, although the best
individual results are now concentrated on indicator
$c_{{\textit{coi}[\aleph,5,\alpha\aleph]}}$.

\paragraph{Overview}

Overall, the results show a remarkable stability in the treatment of
both resource-intensive and resource-scarce languages, the latter
being more dependent on the learning architecture used. The fact that
in most cases the behavior improves for reduced lengths ($k=2$) would
suggest that the configuration $k=5$ used in our previous comparison
with the state-of-the-art is not optimal, thus giving greater value to
the results previously discussed.

\section{Conclusions}
\label{section-conclusions}

We have formally described an early stopping technique to anticipate
overfitting in {\sc dl}-based learning, exploiting cross-validation as
the predictive basis. The novelty compared to previous work lies in
formally tackling two key issues: the specification of early stopping
criteria and the interpretation of agreement among these.  In the
first case, we have described a well-defined framework based on the
notion of the canary function from {\sc gp}, using online indicators
to warn of overfitting.  In the second case, we have analyzed the
agreement in the diagnosis from the perspective of the {\sc pcc}, thus
giving theoretical support to our decision-making.

Our case study focused on the particularly complex task of generating
{\sc nlp} parsers. The results indicate that our proposal improves
generalization capabilities while increasing learning efficiency and
reducing sensitivity with respect to parameter choice. This shows the
potential of correlating decision criteria as a means of providing
stability and preventing overfitting during model training.

\section*{Acknowledgments}
\begin{small}
  Research partially funded by the Spanish Ministry of Economy and
  Competitiveness through projects TIN2017-85160-C2-2-R and
  PID2020-113230RB-C22, and by the Galician Regional Government under
  project ED431C 2018/50. Funding for open access charge: Universidade
  de Vigo/CISUG.
\end{small}

\begin{footnotesize}

\end{footnotesize}

\end{document}